\documentclass[lettersize,journal]{IEEEtran}
\usepackage{amsmath,amsfonts}
\usepackage{algorithmic}
\usepackage{algorithm}
\usepackage{array}

\ifCLASSOPTIONcompsoc
\usepackage[caption=false,font=normalsize,labelfont=sf,textfont=sf]{subfig}
\else
\usepackage[caption=false,font=footnotesize]{subfig}
\fi

\usepackage{textcomp}
\usepackage{stfloats}
\usepackage{hyperref}
\usepackage{url}
\usepackage{verbatim}
\usepackage{graphicx}
\usepackage{color}
\usepackage{cite}
\usepackage{multirow}
\usepackage{tabularx}
\usepackage{ragged2e}
\newcolumntype{P}[1]{>{\Centering\hspace{0pt}}p{#1}}
\newcolumntype{Z}{>{\centering\let\newline\\\arraybackslash\hspace{0pt}}X}
\begin{document}

\title{SCCAM: Supervised Contrastive Convolutional Attention Mechanism for Ante-hoc Interpretable Fault Diagnosis with Limited Fault Samples}
%An Ante-hoc Interpretable Framework via Convolutional Block Attention Module and Supervised Contrastive Learning for Limited Fault Diagnosis}

\author{Mengxuan Li*, Peng Peng*,~\IEEEmembership{Member,~IEEE,} Jingxin Zhang, Hongwei Wang,~\IEEEmembership{Member,~IEEE,} \\and Weiming Shen,~\IEEEmembership{Fellow,~IEEE}
        % <-this % stops a space
%\thanks{This paper was produced by the IEEE Publication Technology Group. They are in Piscataway, NJ.}% <-this % stops a space
%\thanks{Manuscript received April 19, 2021; revised August 16, 2021.}}
\thanks{This work was supported by the National Natural Science Foundation of China under Grant 62276230. (Corresponding authors: Hongwei Wang and Weiming Shen.)}
\thanks{*The two authors contributed equally to this paper.}
\thanks{
Mengxuan Li is with the College of Computer Science and Technology in Zhejiang University, Hangzhou, 310013, China. (E-mail: mengxuanli@intl.zju.edu.cn).

Peng Peng and Hongwei Wang are with Zhejiang University and the University of Illinois Urbana–Champaign Institute, Haining, 314400, China. (E-mail: pengpeng@intl.zju.edu.cn, hongweiwang@intl.zju.edu.cn).

Jingxin Zhang is with the department of Automation, Southeast University, Nanjing, 210096, China. (e-mail: zjx18@tsinghua.org.cn).

Weiming Shen is with the State Key Laboratory of Digital Manufacturing Equipment and Technology, Huazhong University of Science and Technology, Wuhan, 430074, China. (e-mail: wshen@ieee.org).
}
\thanks{This work has been submitted to the IEEE Transactions for possible publication. Copyright may be transferred without notice, after which this version may no longer be accessible.}
}

%\markboth{IEEE transaction on Neural Networks and Learning Systems}%
%Journal of \LaTeX\ Class Files,~Vol.~14, No.~8, %August~2021}%


\maketitle

\begin{abstract}

In real industrial processes, fault diagnosis methods are required to learn from limited fault samples since the procedures are mainly under normal conditions and the faults rarely occur. Although attention mechanisms have become popular in the field of fault diagnosis, the existing attention-based methods are still unsatisfying for the above practical applications. First, pure attention-based architectures like transformers need a large number of fault samples to offset the lack of inductive biases thus performing poorly under limited fault samples. Moreover, the poor fault classification dilemma further leads to the failure of the existing attention-based methods to identify the root causes. To address the aforementioned issues, we innovatively propose a supervised contrastive convolutional attention mechanism (SCCAM) with ante-hoc interpretability, which solves the root cause analysis problem under limited fault samples for the first time. First, accurate classification results are obtained under limited fault samples. More specifically, we integrate the convolutional neural network (CNN) with attention mechanisms to provide strong intrinsic inductive biases of locality and spatial invariance, thereby strengthening the representational power under limited fault samples. In addition, we ulteriorly enhance the classification capability of the SCCAM method under limited fault samples by employing the supervised contrastive learning (SCL) loss. Second, a novel ante-hoc interpretable attention-based architecture is designed to directly obtain the root causes without expert knowledge. The convolutional block attention module (CBAM) is utilized to directly provide feature contribution behind each prediction thus achieving feature-level explanations. The proposed SCCAM method is tested on a continuous stirred tank heater and the Tennessee Eastman industrial process benchmark. Three common fault diagnosis scenarios are covered, including a balanced scenario for additional verification and two scenarios with limited fault samples (\emph{i.e.}, imbalanced scenario and long-tail scenario). The comprehensive results demonstrate that the proposed SCCAM method can achieve better performance compared with the state-of-the-art methods on fault classification and root cause analysis.
\end{abstract}


\section{Introduction}

%\textcolor{blue}{1. Attention-based methods in fault diagnosis. \\ 2. No root cause analysis is provided in the existing attention-based methods. Attention mechanisms have the potential to provide interpretability. Pure attention-based architectures have weak inductive biases thus requiring a mass of data. We utilize CNN to address this issue. \\ 3. We propose SCCAM. For the first time, the root cause analysis problem for limited fault diagnosis is solved. First, an attention-based CNN architecture is designed. More specifically, CBAM is used to capture global information from both channel and spatial domains and directly visualize feature contribution behind each prediction. It can be integrated into any CNN-based architecture with negligible overheads. $1 \times 1$ convolution kernels are used to extract local information and maintain the size of the input data, thus no downsampling or upsampling operations are required. No information is lost and feature-level explanations are enabled. Second, we use SCL loss to extend the proposed SCCAM method to limited fault diagnosis. Powerful feature representations are learned by contrasting samples thus clustering the samples belonging to the same class and distinguishing the samples from different classes. The main contributions of this paper can be summarized as follows.}

\IEEEPARstart{A}{s} THE major part of heavy industry, industrial processes require multi-level hierarchical optimization and control for safe operations while anomalies and faults lead to serious security problems and economic losses \cite{qin2012survey-fd}. Therefore, it is necessary to apply intelligent fault diagnosis methods to detect faults in time and identify the root causes. In recent years, attention mechanisms become increasingly popular in the field of fault diagnosis due to the ability to extract global information and achieve efficient resource allocation \cite{lv2022attention-ifd}. Some achievements have been reported to utilize attention mechanisms to improve the performance of models for fault classification. For example, Wang \emph{et al.} \cite{wang2021feature-tnnls1} proposed a multitask attention module to give feature-level attention to specific tasks. Li \emph{et al.} \cite{li2022variational-tnnls2} proposed a variational attention-based transformer network to mine the association relationships within data. %Li \emph{et al.} \cite{li2022nonlinear-atten2} proposed a unidimensional convolution operation with a self-attention mechanism to enhance the nonlinear representation of process operating conditions. Wang \emph{et al.} \cite{wang2022attention-atten3} proposed an attention-guided joint learning method for equipment condition monitoring. 
Zhou \emph{et al.} \cite{zhou2022exploring-atten1} proposed an industrial process optimization vision transformer (IPO-ViT) to use the global receptive field provided by the self-attention mechanism for fault classification.

%Although it seems that attention mechanisms have achieved success in the field of fault diagnosis, the existing attention-based methods are still unsatisfying for real industrial processes. 
It is worth noting that only limited fault samples can be collected in real industrial processes due to the fact that the industrial processes are mainly under normal conditions and the faults seldom occur in real cases \cite{li2022reweighted-tnnls3}. Therefore, for the existing attention-based fault diagnosis methods, the ability to efficiently leverage the limited fault samples and achieve better generalization is required to provide accurate fault classification results, which are also the foundation of the later root cause analysis.

However, this requirement is challenging for the existing attention-based method due to the lack of inductive biases. Inductive biases refer to initial beliefs or assumptions of models to generalize unseen data. In the lack of inductive biases, attention models have no additional constraints and need a large amount of data to implicitly learn inductive biases \cite{dosovitskiy2020image-vit}, thereby they outperform other methods on large-scale datasets \cite{d2021convit}. In contrast, the models like convolutional neural networks (CNNs) with strong inductive biases have superior generalization ability and better classification performance under limited samples. Therefore, due to the lack of intrinsic inductive biases, pure attention-based architectures like transformers benefit from large datasets but easily fall into overfitting when trained with limited fault samples, thus failing to satisfy the requirements for practical applications. To address this issue, we utilize CNN, which naturally equips with the intrinsic inductive biases of locality and spatial invariance \cite{xu2021vitae}, to enhance the generalization capability and improve the classification performance under limited fault samples. However, directly integrating CNN with attention mechanisms will easily cause the overfitting problem in the long-tail scenario since the complexity and the number of parameters of the original CNN model are increased. To further avoid overfitting, we follow the idea of contrastive learning to efficiently utilize each limited fault sample and learn powerful feature representations in the representation space.

In addition to accurately classifying the fault samples, it is even harder to identify the true root causes under limited fault samples. And the existing attention-based fault diagnosis methods only provide fault classification results without the corresponding root cause analysis. However, attention mechanisms theoretically have the potential to improve interpretability and identify the true root causes while strengthening the representational power of models \cite{wiegreffe2019attention}. Motivated by the success of attention mechanisms in producing visual explanations for image data \cite{selvaraju2017grad-cam}, we design an ante-hoc interpretable attention-based architecture to identify the root causes. More specifically, we utilize $1 \times 1$ convolution kernels to maintain the size of the input data and efficiently extract local information, thereby obtaining the feature maps with unchanged size and enabling feature-level explanations. Then the attention module is applied to further integrate global information and directly visualize the feature contribution behind each prediction. In general, the existing interpretability methods can be roughly divided into two categories: ante-hoc interpretability and post-hoc interpretability. The former refers to directly designing interpretable models by viewing internal model parameters or feature summary statistics while the latter refers to the application of interpretability methods to explain a previously trained model without improvement of the model performance. Contrary to the commonly used ante-hoc interpretable bayesian network-based root cause analysis methods \cite{liu2022optimized-bn} and the existing post-hoc interpretability methods \cite{molnar2020interpretable-cm}, our design allows the proposed model to learn observed data and generate visual explanations simultaneously without any expert knowledge.

In this paper, we innovatively propose a supervised contrastive convolutional attention mechanism (SCCAM) for ante-hoc interpretable fault diagnosis with limited samples. This is the first time that a solution is developed to identify the root causes under extremely limited fault samples. %Contrary to the existing fault diagnosis methods, the proposed SCCAM method can classify the fault samples and identify the corresponding root causes simultaneously while no expert knowledge is required. 
First, we utilize an attention module integrated with CNN to provide intrinsic inductive biases and enable feature-level explanations. More specifically, the convolutional block attention module (CBAM) is applied, since it leverages the strengths of both architectures and CNN can be integrated into it with negligible overheads \cite{woo2018cbam}.
%More specifically, CBAM is utilized to strengthen the representational power of CNN and provide feature contribution behind each prediction. With an intermediate feature map generated by the previous convolution layer, CBAM sequentially infers attention maps along both channel and spatial domains, and then multiplies the attention maps to the input feature map for adaptive feature refinement \cite{woo2018cbam}.  
%In addition, the commonly used convolution layers compress the information and decrease the dimensions of the input. Therefore, to visualize the feature contribution, the attention maps are required to upsample and reshape to original input size. However, the uniqueness of each variable is lost in this process, thus the attention maps cannot provide precise contribution for each feature.
%extract local features and generate smaller feature map, then the attention cannot provide precise contribution for each feature, kernel size larger than 1 will compress information. To address this issue, we utilize $1 \times 1$ convolution kernels to maintain the size of the input data and increase the depth of the model, thus obtaining the feature maps with the unchanged size. It enables us to directly identify the root causes and improves the interpretability of our proposed method. 
Second, the classification capability of the proposed SCCAM method is strengthened under limited fault samples by employing the supervised contrastive learning (SCL) loss. SCL is a powerful feature representation learning technique \cite{khosla2020scl}. It not only contrasts samples against each other to learn attributes like traditional self-supervised contrastive learning, but also effectively leverages label information like supervised learning. We apply the SCL loss to cluster the samples belonging to the same category and separate the samples from different classes. The main contributions of this paper can be summarized as follows:
\begin{enumerate}
    \item We propose a novel ante-hoc interpretable fault diagnosis approach under limited fault samples, namely SCCAM. Since CBAM combines the advantages of both CNN and attention mechanisms, it is utilized to enhance the feature extraction capability and provide intrinsic inductive biases thus improving the performance of the attention-based architecture under limited fault samples.
    \item We apply SCL to further improve the classification performance of the proposed SCCAM method under limited fault samples. Samples with the same label are clustered while samples from different classes are disaggregated. It enables the proposed SCCAM method to efficiently utilize each limited fault sample and learn powerful feature representations in the representation space.
    \item We design an innovative ante-hoc interpretable attention-based architecture by combining $1 \times 1$ convolution kernels and CBAM. These kernels are used to maintain the size of the input data and enable feature-level explanations. And CBAM is utilized to directly visualize the feature contribution behind each prediction, thereby we automatically obtain the corresponding root causes without any expert knowledge.
    \item The comprehensive performance of the proposed SCCAM method is analyzed with a continuous stirred tank heater (CSTH) and the Tennessee Eastman (TE) industrial process benchmark. Moreover, three common fault diagnosis scenarios are tested, including a balanced scenario for additional verification and two scenarios with limited fault samples (\emph{i.e.}, imbalanced scenario and long-tail scenario). Experimental results show that the proposed SCCAM method achieves higher fault classification accuracy and better root cause identification compared with the state-of-the-art methods on fault classification and root cause analysis.
\end{enumerate}

The remainder of this paper is structured as follows. Section \ref{prep} briefly introduces the relevant preparations. The SCCAM method proposed in this paper is described in Section \ref{method}. Then the comprehensive experiments are presented in Section \ref{experiment}. Section \ref{conclusion} concludes this paper.

%1. importance of fault diagnosis, data-driven method, deep learning method, CNN
 
%2. need data, not interpretable

%3. interpretability, post-hoc CAM/Grad-CAM not precise; shap not include prior knowledge, ante-hoc ViT multi-head attention not stable, CBAM attention based on CNN

%4. our proposed method

%5. contribution

\section{Preliminary}
\label{prep}

\begin{figure*}[]
\centering
\subfloat[]
{
 \centering
 \label{cbam}
 \includegraphics[width=5.5cm]{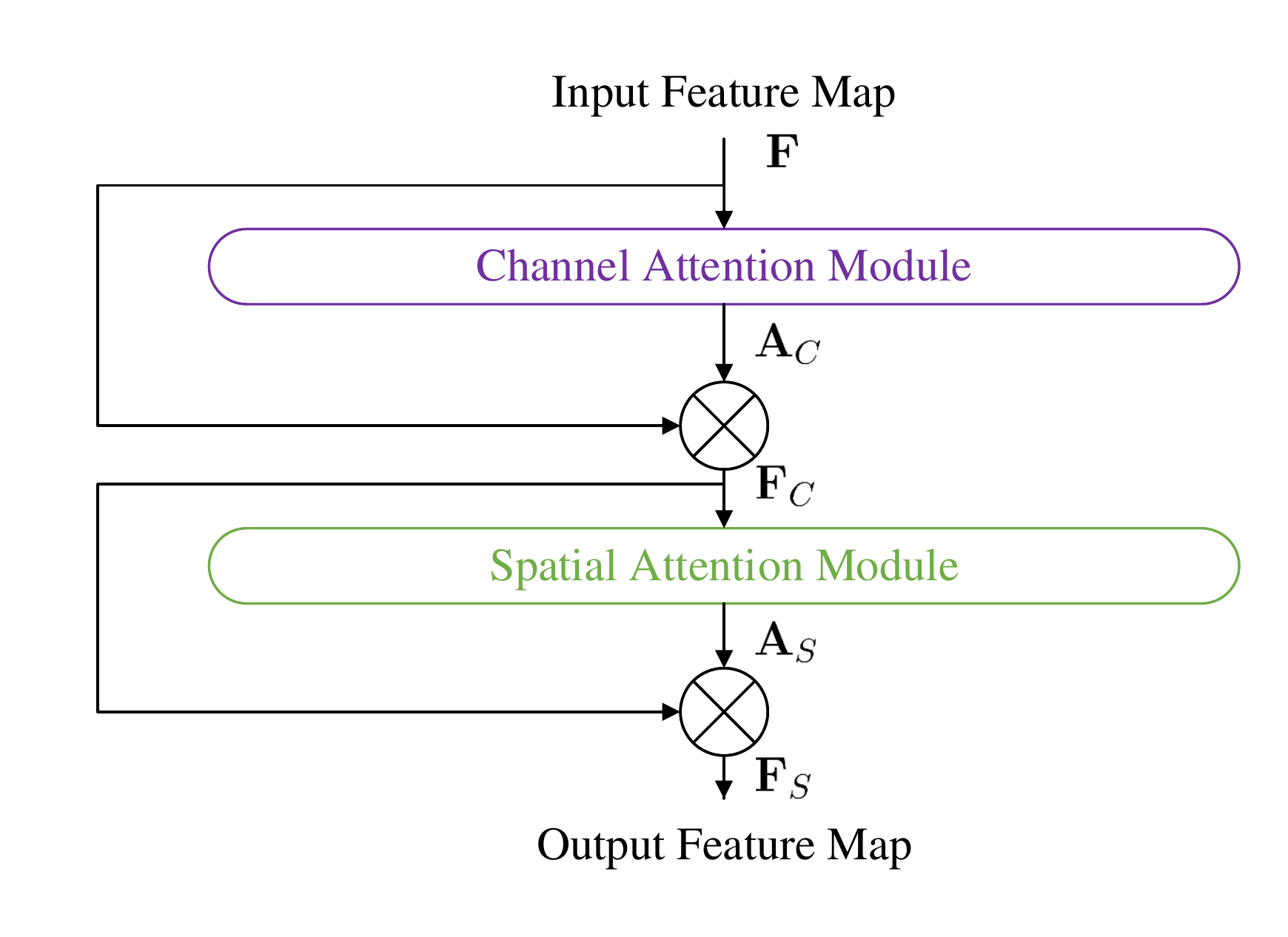}
}
\subfloat[]
{
 \centering
 \label{cbam_c}
 \includegraphics[width=5.5cm]{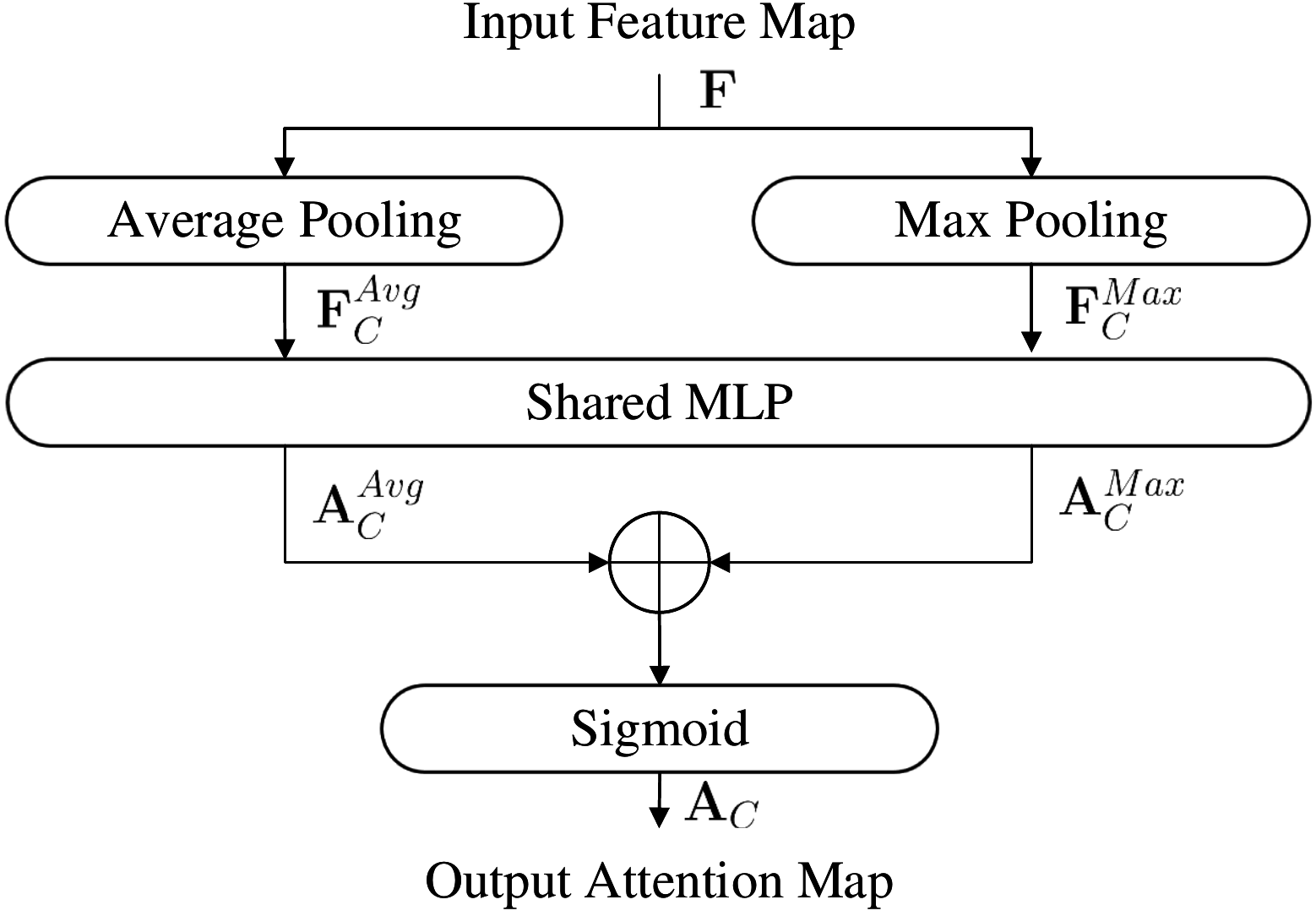}
}
\subfloat[]
{
 \centering
 \label{cbam_s}
 \includegraphics[width=5.5cm]{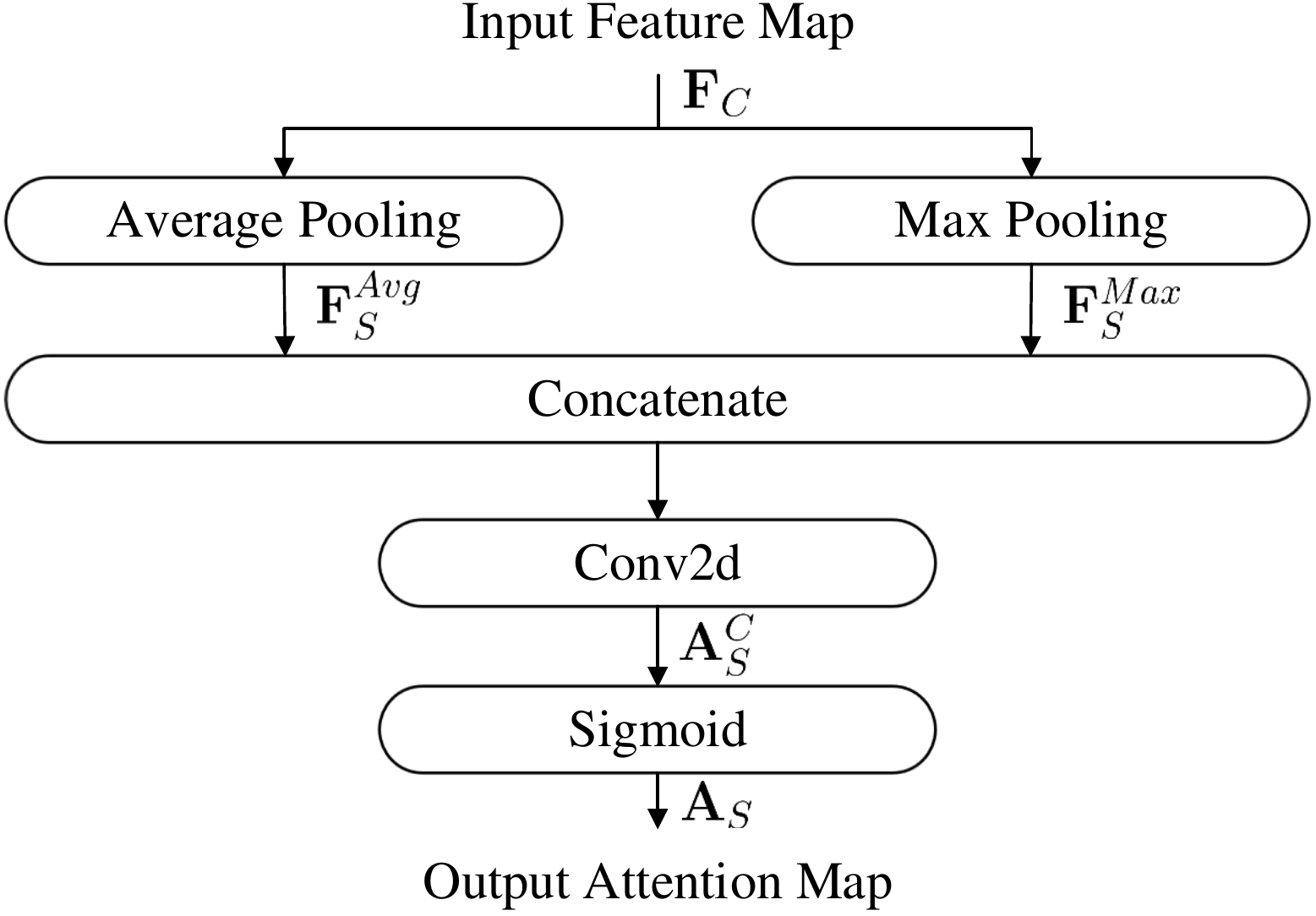}
}
\caption{Architecture of (a) CBAM, (b) channel attention module and (c) spatial attention module.} 
\label{cbam_ar}
\end{figure*}

\subsection{Convolutional Block Attention Module}
\label{cbam_theo}
The attention mechanism \cite{vaswani2017attention} is proposed to highlight the vital features and diminish the rest non-significant parts. It generates weights of the input features, making the deep learning networks devote more focus to the important parts. Furthermore, by visualizing the feature contribution for classification, the attention mechanism can improve interpretability and the root causes can be identified directly. On the basis of it, CBAM \cite{woo2018cbam} is designed to improve the performance of CNN models by integrating the attention mechanism. It generates attention maps along channel dimension and spatial dimension in sequence, and can be incorporated into any CNN framework with negligible overheads \cite{woo2018cbam}. As shown in Fig. \ref{cbam}, for a feature map $\mathbf{F} \in \mathbb{R}^{C \times H \times W}$ extracted by any convolutional layers, a channel attention map $\mathbf{A}_{C} \in \mathbb{R}^{C \times 1 \times 1}$ and a spatial attention map $\mathbf{A}_{S} \in \mathbb{R}^{1 \times H \times W}$ are generated sequentially, where $C$, $H$, $W$ refer to the number of channels, the input height, the input width respectively. Theoretically, the overall algorithm can be identified as:
\begin{align}
\label{cbam2}
\mathbf{F}_{C} &=\mathbf{A}_{C}(\mathbf{F}) \otimes \mathbf{F}\\
\mathbf{F}_{S} &=\mathbf{A}_{S}\left(\mathbf{F}_{C}\right) \otimes \mathbf{F}_{C}
\label{cbam1}
\end{align}
where $\otimes$ refers to element-wise multiplication, $\mathbf{F}_{C}$ denotes the feature map refined by channel attention values, and $\mathbf{F}_{S}$ is the final output feature map refined by both channel and spatial attention values. The details of the channel attention module and spatial attention module are demonstrated in Fig. \ref{cbam_c} and Fig. \ref{cbam_s}.

\subsubsection{Channel Attention Module}
Firstly, the channel attention module is designed to extract important features among channels. Average-pooling and max-pooling operations are utilized simultaneously to gather spatial information, thus allowing us to fully focus on the channel information. Average-pooled feature map $\mathbf{F}_{C}^{Avg} \in \mathbb{R}^{C \times 1 \times 1}$ and max-pooled feature map $\mathbf{F}_{C}^{Max} \in \mathbb{R}^{C \times 1 \times 1}$ are generated as follows:
\begin{align}
    \mathbf{F}_{C_i}^{Avg} &= \frac{1}{H*W}\sum_{m=0}^{H-1}\sum_{n=0}^{W-1}\mathbf{F}(C_i,m,n)\\
    \mathbf{F}_{C_i}^{Max} &= \max \limits_{m=0,...,H-1}\max \limits_{n=0,...,W-1} \mathbf{F}(C_i,m,n)
\end{align}
where $C_i$ refers to the $i$-th channel. Then, the two pooled feature maps are used to compute the channel attention map $\mathbf{A}_{C}$ with a shared multi-layer perceptron (MLP). The shared MLP has three layers: a) a fully connected layer to transform the dimensions of the input feature maps from $\mathbb{R}^{C \times 1 \times 1}$ to $\mathbb{R}^{C/r \times 1 \times 1}$, where $r$ is the reduction ratio to decrease the parameter costs; b) a hidden layer with ReLU activation function to complete the nonlinear transformation of data and alleviate the overfitting problem; c) a fully connected layer to convert the dimensions back to $\mathbb{R}^{C \times 1 \times 1}$. The process can be calculated as follows:
\begin{align}
    \mathbf{A}_{C}^{Avg} &= \mathbf{W_1}(\sigma(\mathbf{W_0}(\mathbf{F}_{C}^{Avg})))\\
    \mathbf{A}_{C}^{Max} &= \mathbf{W_1}(\sigma(\mathbf{W_0}(\mathbf{F}_{C}^{Max})))
\end{align}
where $\sigma$ denotes the ReLU activation function, $\mathbf{W_0} \in \mathbb{R}^{C / r \times C}$ and $\mathbf{W_1} \in \mathbb{R}^{C \times C / r}$ refer to the learnable weights of the two fully connected layers. The channel attention map $\mathbf{A}_{C}$ is then computed by adding $\mathbf{A}_{C}^{Avg}$ and $\mathbf{A}_{C}^{Max}$. In addition, a sigmoid activation function is utilized for normalization:
\begin{align}
    \mathbf{A}_{C} &= \frac{1}{1+\exp (-(\mathbf{A}_{C}^{Avg} + \mathbf{A}_{C}^{Max}))}.
\end{align}
Then, the new feature map $\mathbf{F}_{C}$ is obtained by multiplying the channel attention values $\mathbf{A}_{C}$ and the original feature map $\mathbf{F}$.

\subsubsection{Spatial Attention Module}
As a complement to channel attention, the spatial attention module is designed to focus on the vital area of the input features. Similarly, average-pooled feature map $\mathbf{F}_{S}^{Avg} \in \mathbb{R}^{1 \times H \times W}$ and max-pooled feature map $\mathbf{F}_{S}^{Max} \in \mathbb{R}^{1 \times H \times W}$ are generated to integrate the channel information:
\begin{align}
    \mathbf{F}_{S}^{Avg} &= \frac{1}{C}\sum_{k=0}^{C-1}\mathbf{F}(k,H,W)\\
    \mathbf{F}_{S}^{Max} &= \max \limits_{k=0,...,C-1} \mathbf{F}(k,H,W).
\end{align}
Those feature maps are concatenated to obtain the feature map cross channels and then forwarded to a convolutional layer to further extract spatial attention map $\mathbf{A}_{S}^{C} \in \mathbb{R}^{1 \times H \times W}$, which is normalized with a sigmoid activation function to obtain the spatial attention map $\mathbf{A}_{S}$:
\begin{align}
    \mathbf{A}_{S}^{C} &= f^{\alpha \times \alpha}([\mathbf{F}_{S}^{Avg};\mathbf{F}_{S}^{Max}])\\
    \mathbf{A}_{S} &= \frac{1}{1+\exp (-\mathbf{A}_{S}^{C})}
\end{align}
where $f^{\alpha \times \alpha}$ denotes a convolution operation and the filter size is $\alpha \times \alpha$. Finally, the output feature map $\mathbf{F}_{S}$ is computed as shown in Equation \ref{cbam1}.

%%%%%%%%%%%%%%%%%%%%%%%%%
\begin{figure*}[]
\centering
\includegraphics[width = \textwidth]{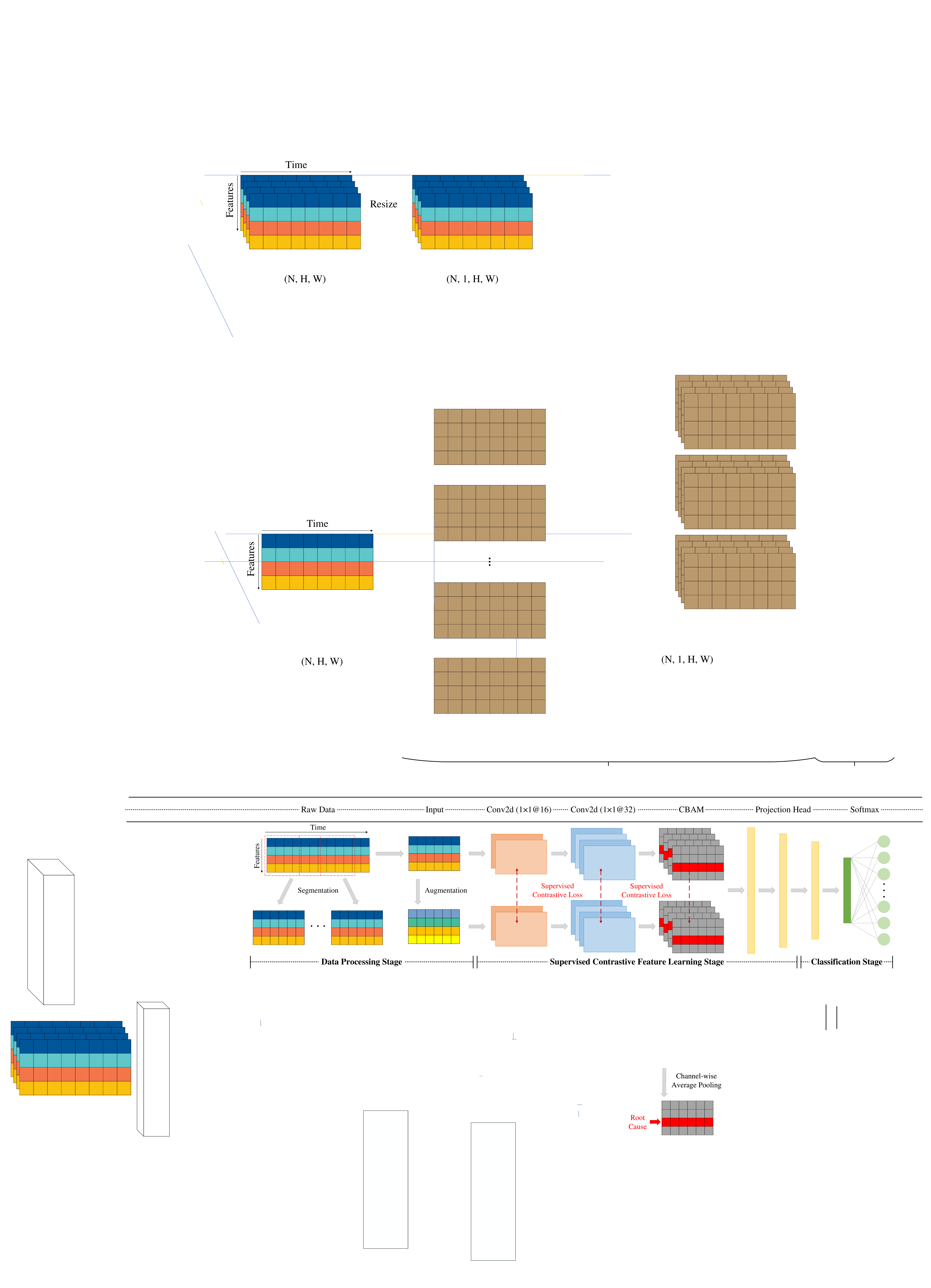}
\caption{The flow of the proposed SCCAM model. Red regions refer to higher scores in attention maps (\emph{i.e.}, higher feature contribution).}
\label{fig_network}
\end{figure*}

\subsection{Supervised Contrastive Learning}
Generally, contrastive learning is considered as one of the most powerful approaches to self-supervised learning that learns feature representation without label information by clustering similar samples and distinguishing dissimilar ones. As an extension of self-supervised contrast learning, SCL is robust and stable due to the ability to leverage label information effectively to generate augmented samples \cite{khosla2020scl}. For a batch with $N$ randomly sampled data $\left\{\boldsymbol{x}_k, {y}_k\right\}_{k=1 \ldots N}$, the corresponding augmented batch consists of $2N$ generated data $\left\{\tilde{\boldsymbol{x}}_{\ell}, \tilde{{y}}_{\ell}\right\}_{\ell=1 \ldots 2 N}$. $\tilde{\boldsymbol{x}}_{2 k}$ and $\tilde{\boldsymbol{x}}_{2 k-1}$ are two random augmentations of $\tilde{\boldsymbol{x}}_{k}$, and  $\tilde{{y}}_{2k-1} = \tilde{{y}}_{2k} = \tilde{{y}}_{k}$. The computational details of the corresponding self-supervised contrastive loss and SCL loss are demonstrated in the following parts.

\subsubsection{Self-Supervised Contrastive Loss}
Self-supervised contrastive learning is typically applied to pull together the representations of an "anchor" and a "positive" sample while pushing apart this "anchor" from other "negative" samples. In this case, for an anchor $\tilde{\boldsymbol{x}}_{i}$, define the other augmented sample generating from the same original data as $\tilde{\boldsymbol{x}}_j(i)$. Then $\tilde{\boldsymbol{x}}_j(i)$ is called positive, and the rest $(2N-2)$ augmented samples in $\left\{\tilde{\boldsymbol{x}}_{\ell}, \tilde{{y}}_{\ell}\right\}_{\ell=1 \ldots 2 N}$ are called negatives. The self-supervised contrastive loss is defined as follows:
\begin{equation}
\label{cl}
\mathcal{L}^{\text {self }}=\sum_{i \in I} \mathcal{L}_i^{\text {self }}=-\sum_{i \in I} \log \frac{\exp \left(\boldsymbol{z}_i \cdot \boldsymbol{z}_{j(i)} / \tau\right)}{\sum_{a \in A(i)} \exp \left(\boldsymbol{z}_i \cdot \boldsymbol{z}_a / \tau\right)}
\end{equation}
where $i \in I \equiv\{1 \ldots 2N\}$ is the index of an arbitrary anchor, $\boldsymbol{z}_i$ refers to the embedding of $\tilde{\boldsymbol{x}_i}$ extracted by the corresponding contrastive network, $\tau \in \mathbb{R}^{+}$ is a scalar temperature parameter, and $A(i) \equiv I \backslash\{i\}$. In this equation, the denominator has $(2N-1)$ terms in total including the positive and negatives.

\subsubsection{Supervised Contrastive Loss}
With supervised data, the contrastive loss shown in Equation \ref{cl} is inappropriate since a class contains multiple known samples. In this case, for an anchor $\tilde{\boldsymbol{x}}_{i}$, the other augmented samples generated from the data belonging to the same class are called positives, and the remaining augmented samples in $\left\{\tilde{\boldsymbol{x}}_{\ell}, \tilde{{y}}_{\ell}\right\}_{\ell=1 \ldots 2 N}$ are called negatives. The SCL loss is computed as follows:
\begin{align}
\mathcal{L}^{\text {sup }} &=\sum_{i \in I} \mathcal{L}_{i}^{\text {sup }}\nonumber\\
&=\sum_{i \in I} \frac{-1}{|P(i)|} \sum_{p \in P(i)} \log \frac{\exp \left(\boldsymbol{z}_i \cdot \boldsymbol{z}_p / \tau\right)}{\sum_{a \in A(i)} \exp \left(\boldsymbol{z}_i \cdot \boldsymbol{z}_a / \tau\right)}
\label{scl_loss}
\end{align}
where $P(i) \equiv\left\{p \in A(i): \tilde{{y}}_p=\tilde{{y}}_i\right\}$ refers to the set of indices of all positives of the anchor $\tilde{\boldsymbol{x}}_{i}$, and $|P(i)|$ is its cardinality.

\section{Methodology}
\label{method}

%\subsection{Overall Framework of the proposed SCCAM method}
In this paper, we innovatively propose SCCAM, an ante-hoc interpretable fault diagnosis method under limited fault samples. The core idea is to utilize CBAM to not only enhance the local feature extraction capability and provide intrinsic inductive biases, but also capture global information and visualize feature contribution. And the SCL loss is applied to efficiently leverage each limited fault sample and learn powerful representations. As shown in Fig. \ref{fig_network}, the proposed model mainly includes three stages: data processing stage, supervised contrastive feature learning stage, and classification stage. When the offline model is trained, the online fault diagnosis can be performed as shown in Fig. \ref{fig_fault_diagnosis}. The details are introduced in the following parts.

\subsection{Data Processing Stage}
This stage aims to obtain standardized and augmented input from the raw data for model training. Since the collected data in industrial processes is noisy and variable in the value range, the features are standardized to reduce the effect of inconsistent data on model performance \cite{liu2022fusion-tcy}. The standardized data $\hat{\boldsymbol{X}}$ then has a mean of 0 and standard deviation of 1 by removing the mean and scaling to unit variance:
\begin{equation}
\hat{\boldsymbol{X}}=\frac{\boldsymbol{X}-\bar{\boldsymbol{X}}}{{\boldsymbol{S}}}
\end{equation}
where $\bar{\boldsymbol{X}}$ and $\boldsymbol{S}$ refer to the mean and the standard deviation of the original data. Then inspired by \cite{chen2021graph-tcy}, we use the sliding-window method to segment the long time-series data into multiple shorter data for later analysis. For each fault type ${c}$, the standardized time-series data is sliced to produce the corresponding dataset $\hat{\boldsymbol{X}}_{c}=\{\hat{\boldsymbol{x}}_{1},\hat{\boldsymbol{x}}_{2},\ldots,\hat{\boldsymbol{x}}_{N}\} \in \mathbb{R}^{N \times H \times W}$, which means the dataset includes $N$ samples, each sample has $H$ features, and each feature is measured within time $W$. With the preprocessed dataset, a random augmentation $\tilde{\boldsymbol{x}}_{i}$ is generated for each data sample $\hat{\boldsymbol{x}}_{i}$ as follows:
\begin{equation}
    \tilde{\boldsymbol{x}}_{i} = \hat{\boldsymbol{x}}_{i} + noise
\end{equation}
where the noise matrix is filled with random noise generated from a standard normal distribution:
\begin{equation}
    noise_{j} \sim \mathcal{N}(0,1).
\end{equation}
Finally, for each fault type, we obtain a standardized and augmented dataset composed by $2N$ samples $\left\{\tilde{\boldsymbol{x}}_{\ell}, \tilde{{y}}_{\ell}\right\}_{\ell=1 \ldots 2 N}$, where $\tilde{\boldsymbol{x}}_{2k-1}=\hat{\boldsymbol{x}}_{k}$, $\tilde{\boldsymbol{x}}_{2k}$ is the random augmentation of $\hat{\boldsymbol{x}}_{k}$, and $\tilde{{y}}_{2 k-1} = \tilde{{y}}_{2k} = \tilde{{y}}_{k}$.

\subsection{Supervised Contrastive Feature Learning Stage}
In this stage, the SCL loss is applied to learn powerful feature representations from the limited data. More specifically, an encoder network is trained to map the input $\tilde{\boldsymbol{x}}_{\ell}$ to a representation vector $\boldsymbol{z}_{\ell}$ with the SCL loss defined in Equation \ref{scl_loss}: $\boldsymbol{z}_{\ell}=Encoder(\tilde{\boldsymbol{x}}_{\ell})$. As shown in Fig. \ref{fig_network}, the encoder network consists of a two-layer convolutional network, a CBAM module, and a projection head. Details are described below.

\subsubsection{Two-layer Convolutional Network}
The two-layer convolutional network is designed to learn complex features from the input data. Each layer applies a convolution operation to extract the corresponding feature map. The convolution operation at the ($l$ + 1)-th
layer can be expressed as follows:
\begin{equation}
\tilde{\boldsymbol{x}}_{\ell}^{l+1}=f^{1 \times 1}(\tilde{\boldsymbol{x}}_{\ell}^{l} , {\boldsymbol{K}}^{l+1}) +{\boldsymbol{b}}_{\ell}^{l+1}
\end{equation}
where $\tilde{\boldsymbol{x}}_{\ell}^{l}$ is the input of the ($l$ + 1)-th layer, ${\boldsymbol{K}}^{l+1}$ is the convolution kernel at the ($l$ + 1)-th layer, ${\boldsymbol{b}}_{\ell}^{l+1}$ is the corresponding bias at the ($l$ + 1)-th layer, $f^{1 \times 1}$ denotes a convolution operation and the filter size is $1 \times 1$. Then we apply batch normalization operation to standardize the input to a layer for each mini-batch, thus handling the problem of internal covariate shift. Assume $\tilde{\boldsymbol{x}}_{\ell}$ belongs to a mini-batch $\tilde{\boldsymbol{X}}^{*}$ of size $B$, the empirical mean and variance can be calculated as follows:
\begin{align}
    \boldsymbol{\mu}&=\frac{1}{B} \sum_{i=1}^B \tilde{\boldsymbol{x}}^{*}_{i} \\ {\sigma}^2&=\frac{1}{B} \sum_{i=1}^B\left(\tilde{\boldsymbol{x}}^{*}_i-\boldsymbol{\mu}\right)^2.
\end{align}
The corresponding batch normalization operation is defined as follows:
\begin{equation}
    \tilde{\boldsymbol{x}}_{\ell}^{*}=\frac{{\tilde{\boldsymbol{x}}_{\ell}^{l+1}}-\boldsymbol{\mu}}{\sqrt{\sigma^2+\epsilon}} * \gamma + \beta
\end{equation}
where $\epsilon$ is an arbitrary small constant for numerical stability, $\gamma$ and $\beta$ are learnable parameters. And a ReLU activation function is appiled to execute nonlinear transformation and mitigate the vanishing gradient problem. The expression is as follows:
\begin{equation}
    \boldsymbol{F}_{\ell}^{l+1} = \max(0,\tilde{\boldsymbol{x}}_{\ell}^{*}).
\end{equation}
where $\boldsymbol{F}_{\ell}^{l+1}$ denotes the feature map at the ($l$ + 1)-th layer.

As shown in Fig. \ref{fig_network}, the two convolutional layers have $16$ and $32$ kernels respectively, each of which is with the size of $1 \times 1$. These kernels can increase the depth of the network without changing the size of feature maps or losing resolution information, thus greatly enlarging the non-linearity and improving the feature extraction ability of the whole network. The $1 \times 1$ convolution offers filter-wise pooling, acting as a projection layer to increase the number of channels and extract features across channels. More importantly, each channel of the output feature map of the $1 \times 1$ convolutional layer is identical to the input data in size, resulting in the ability to compute the feature contribution from the obtained attention map directly. In other words, the $1 \times 1$ convolution kernels offer ante-hoc interpretability and enable feature-level explanations without any expert knowledge. Finally, the feature map ${\boldsymbol{F}}_{\ell} \in \mathbb{R}^{32 \times H \times W}$ is obtained. 
%In addition, each convolutional layer is stacked with a batch normalization layer and a ReLU activation function.

\subsubsection{CBAM Module}
The attention-based CBAM module is utilized to further improve the feature extraction ability and compute the feature contribution behind each prediction. More specifically, channel attention and spatial attention are applied sequentially to integrate global and local information. On the one hand, channel attention globally exploits the inter-channel relationship of features while ignoring the local information within each channel. On the other hand, spatial attention locally concentrates on domain space encapsulated within each feature map whilst neglecting the global information across channels. Therefore, combining both can integrate global and local information simultaneously, thus robustly enhancing the model performance. With the input feature map ${\boldsymbol{F}}_{\ell} \in \mathbb{R}^{32 \times H \times W}$, the output feature map ${\boldsymbol{F}}_{\ell}^{S} \in \mathbb{R}^{32 \times H \times W}$ is obtained:
\begin{equation}
    {\boldsymbol{F}}_{\ell}^{S} = CBAM({\boldsymbol{F}}_{\ell})
\end{equation}
where $CBAM(\cdot)$ refers to the computational method mentioned in Equation \ref{cbam2} and \ref{cbam1}. The CBAM module makes our model focus on important features and ${\boldsymbol{F}}_{\ell}^{S}$ is the final refined output. We generate heatmap visualization of ${\boldsymbol{F}}_{\ell}^{S}$ to show the feature contribution and identify the root causes.

\subsubsection{Projection Head}
Inspired by \cite{chen2020simple-simclr}, a projection head is added to improve the representation quality of the network, which maps representations to the space where SCL loss is applied. Theoretically, the feature map ${\boldsymbol{F}}_{\ell}^{S} \in \mathbb{R}^{32 \times H \times W}$ is firstly flattened to obtain the 1-D feature ${\boldsymbol{h}}_{\ell}$:
\begin{equation}
    {\boldsymbol{h}}_{\ell} = flatten({\boldsymbol{F}}_{\ell}^{S}).
\end{equation}
Since ${\boldsymbol{h}}_{\ell}$ retains the information related to the augmented data, the non-linear projection head is utilized to remove such information and represent the original features. The flattened vector ${\boldsymbol{h}}_{\ell}$ is forwarded into the projection head, which is designed as an MLP with one hidden layer:
\begin{align}
    {\boldsymbol{h}}_{\ell}^{\prime} &= \max(0,\boldsymbol{W}^{1} ({\boldsymbol{h}}_{\ell}))\\
    {\boldsymbol{z}}_{\ell} &= \boldsymbol{W}^{2}({\boldsymbol{h}}_{\ell}^{\prime})
\end{align}
where $\boldsymbol{W}^{1}$ and $\boldsymbol{W}^{2}$ are learnable weights. At this point, we obtain the representation vector $\boldsymbol{z}_{\ell}$ with the SCL loss.

%%%%%%%%%%%%%%%%%%%%%%%%%%%%%%%%%%%%%%%%%%
\begin{figure*}[]
\centering
\includegraphics[width = \textwidth]{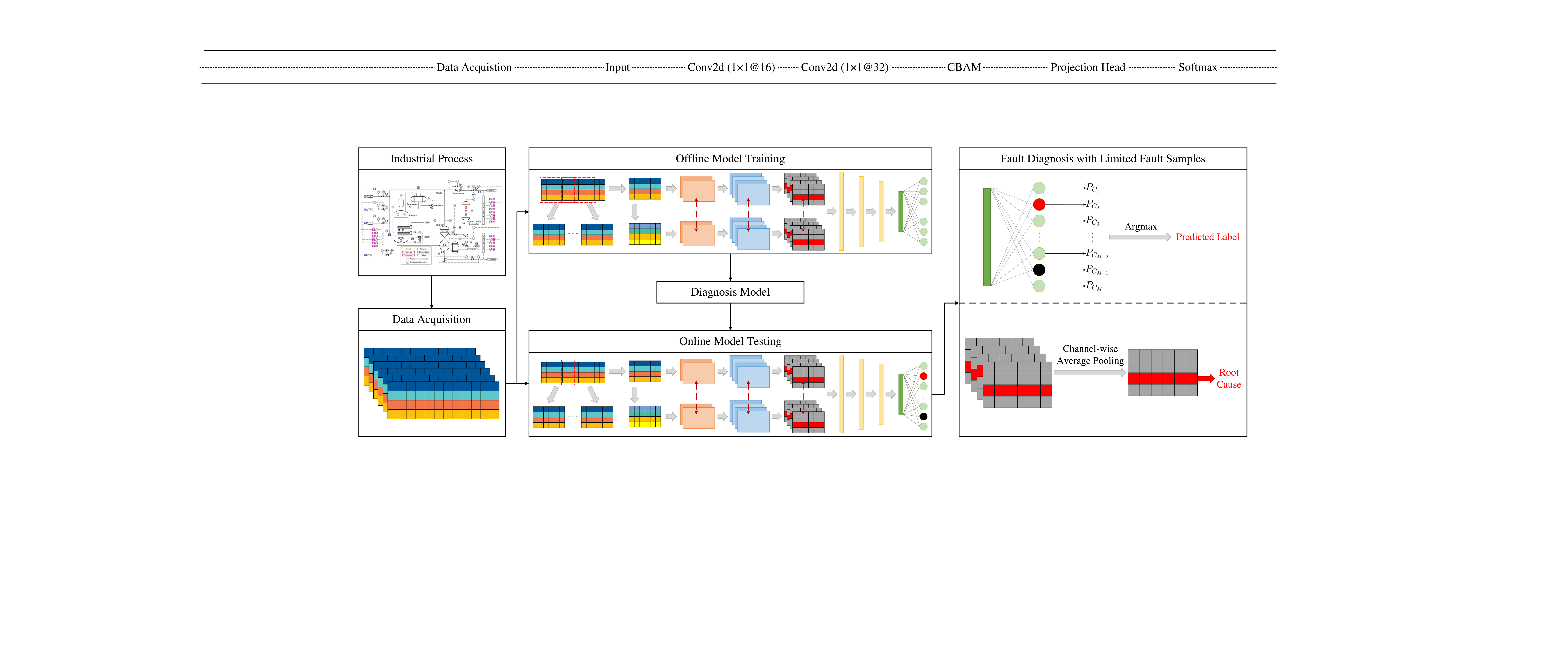}
\caption{The flowchart of the proposed SCCAM method with ante-hoc interpretability for fault diagnosis under limited fault samples. The industrial process is an example of the TE process from \cite{ma2020novel-teimg}.}
\label{fig_fault_diagnosis}
\end{figure*}

\subsection{Classification Stage}
The purpose of this stage is to classify the learned representation vector $\boldsymbol{z}_{\ell}$ into ${M}$ fault types. A fully connected layer is applied to map the feature representation to the output class labels:
\begin{equation}
    \boldsymbol{p}_{\ell} = \boldsymbol{W}(\boldsymbol{z}_{\ell})
\end{equation}
where $\boldsymbol{p}_{\ell} \in \mathbb{R}^{{M}}$ and $\boldsymbol{W}$ is the weight matrix. Then the conditional probability of each class is computed with the softmax function:
\begin{equation}
    P(c \mid \boldsymbol{p}_{\ell}) = \frac{\exp \left( \boldsymbol{p}_{\ell,c}\right)}{\sum_{i=1}^{{M}} \exp \left(\boldsymbol{p}_{\ell,i}\right)}
\end{equation}
where $c$ refers to the class set. And the predicted fault label $\tilde{y}_{\ell}$ is defined as the one with the highest probability:
\begin{equation}
    \tilde{y}_{\ell} = \underset{c}{\arg \max}(P(c \mid \boldsymbol{p}_{\ell})).
\end{equation}
Unlike the previous stage, the cross-entropy (CE) loss is applied to adjust the model weights and optimize the classification. The calculation is defined as follows:
\begin{equation}
    \mathcal{L}^{\text {ce}} = -\sum^{M}_{c=1}y_{\ell}\log(P(c \mid \boldsymbol{p}_{\ell}))
\end{equation}
where $y_{\ell}$ is the real target.

\subsection{Fault Diagnosis Procedure}

Fig. \ref{fig_fault_diagnosis} illustrates the flowchart of the proposed ante-hoc interpretability fault diagnosis method. The fault diagnosis procedure consists of two steps: offline model training and online model testing. The first step aims to obtain a trained model for fault diagnosis under limited fault samples. The second step is designed to classify the fault type and identify the corresponding root cause of the input data.

\subsubsection{Offline Model Training}
In this step, the monitoring values of sensors are collected from the industrial process. Then the acquired data is fed into the offline training model described in Fig. \ref{fig_network}. More specifically, the training samples are obtained from the collected data in the data processing stage. Then the training set is utilized to obtain a powerful encoder based on the SCL loss in the supervised contrastive feature learning stage. A robust classifier is trained based on the cross-entropy loss in the following classification stage. Finally, we obtain a trained model including an encoder and a linear classifier.

\subsubsection{Online Model Test}
In this step, the online data is acquired and processed as test samples. Based on the offline model trained in the previous step, each test sample is classified and a predicted label is obtained through the softmax function. If the test sample belongs to a fault class, the attention map generated by CBAM is utilized to identify the corresponding root cause. As shown in Fig. \ref{fig_fault_diagnosis}, a channel-wise average pooling operation is applied to integrate information across channels and an attention map is generated to denote the feature contribution. Finally, the root cause is identified as the variable with the highest feature contribution.

\section{Case Study}
\label{experiment}

In this paper, a typical CSTH process system with 5 variables is used to develop and test the proposed SCCAM method. And the TE benchmark dataset with 22 variables is applied for further comprehensive evaluation and analysis, including an inductive biases discussion and ablation experiments. We compare our proposed SCCAM method with four state-of-the-art long-tail learning algorithms: focal loss \cite{lin2017focal}, seesaw loss \cite{wang2021seesaw},  label-distribution-aware margin (LDAM) loss \cite{cao2019learning-ldam}, and progressively balanced supervised contrastive learning (PBS-SCL) \cite{peng2022progressively}.
%\cite{zhou2022exploring-atten1}: vision transformer (ViT) \cite{dosovitskiy2020image-vit}, CNN \cite{wang2020intelligent-tcy}, and a simple combination of CNN and CBAM (CNN\_CBAM). 
%ViT is a popular ante-hoc interpretability method that utilizes multi-head attention to extract feature representations \cite{dosovitskiy2020image-vit}. CNN is the baseline and CNN\_CBAM is trained similarly to the proposed SCCAM method but with a traditional cross-entropy loss.
%CNN, ViT \cite{dosovitskiy2020image-vit} and a combination of CNN and CBAM. CNN is the baseline of the proposed SCCAM method and ViT is a popular ante-hoc interpretability method which utilizes multi-head attention to extract feature representations. The combination of CNN and CBAM (CNN\_CBAM) means the model is trained similar to the proposed SCCAM method but with a traditional cross-entropy loss. CNN\_CBAM\_SCL is the abbreviation of our proposed method. 
Comprehensive experiments are conducted under three common fault diagnosis scenarios to verify the classification capability of the proposed SCCAM method, involving a balanced scenario for additional verification and two scenarios with limited fault samples (\emph{i.e.}, imbalanced scenario and long-tail scenario). And the root cause analysis is demonstrated in the long-tail scenario. The comparison results show the effectiveness of our proposed method in classifying the fault samples and identifying the root cause. Details are described in the following parts. %\textcolor{red}{ablation experiments explanation - CNN/CNN\_CBAM/CNN\_CBAM\_SCL}

\subsection{Case 1: Continuous Stirred Tank Heater}

\subsubsection{Brief Introduction of CSTH dataset}

%%%%%%%%%%%%%%%%%%%%%%%%%%%%%%%%%%%%%%%%%%%%%%%%%%%%%%
\begin{comment}
\begin{figure}[b]
\centering
\includegraphics[width = 6cm]{csth.pdf}
\caption{Process flowchart of the CSTH process \cite{thornhill2008continuous-csth}.}
\label{fig_csth_arch}
\end{figure}
\end{comment}

\begin{table}[b]
\footnotesize
\centering
\caption{Process Scenarios for CSTH Dataset \cite{arunthavanathan2022autonomous-csthex}}
%\resizebox{\linewidth}{!}{
\begin{tabularx}{\linewidth}{P{4cm}Z}
\hline
Fault ID & Description           \\ \hline
0        & Normal case           \\
1        & Steam valve position \\ \hline
\end{tabularx}
\label{table_csth_fault}
\end{table}

The CSTH is a typical pilot plant for control performance evaluation and process monitoring, which is non-linear and contains real disturbance data. %As shown in Fig. \ref{fig_csth_arch}, 
More detailedly, hot and cold water are mixed and heated using steam through a heating coil. Then the processed water is drained from the tank through a long pipe \cite{thornhill2008continuous-csth}. 5 variables are involved and two conditions described in Table \ref{table_csth_fault} are considered to test the effectiveness of our proposed SCCAM method. In the Fault 1 condition, a fault is applied in the steam valve position, thus the steam value is the corresponding root cause \cite{arunthavanathan2022autonomous-csthex}.

\begin{table}[]
\caption{Experimental Settings on CSTH Dataset}
\centering
\resizebox{\linewidth}{!}{
\begin{tabular}{ccccccc}
\hline
\multirow{2}{*}{Dataset} & \multicolumn{2}{c}{Balance} & \multicolumn{2}{c}{Imbalance} & \multicolumn{2}{c}{Long-tail} \\ \cline{2-7} 
                         & Normal        & Fault       & Normal         & Fault        & Normal        & Fault        \\ \hline
Train                    & 780          & 450        & 780           & 200          & 780          & 20           \\
Test                     & 200           & 200         & 200            & 200          & 200           & 200          \\ \hline
\end{tabular}}
\label{tabel_csth_setting}
\end{table}

\begin{table}[]
\caption{Comparison of Fault Diagnosis Results on CSTH Dataset}
\centering
\setlength{\tabcolsep}{4.6mm}{
\begin{tabular}{cccc}
\hline
Method      & Balance         & Imbalance       & Long-tail       \\ \hline
Focal Loss \cite{lin2017focal}  & 99.43           & 98.57           & 98.00           \\
Seesaw Loss \cite{wang2021seesaw} & 99.51           & 98.97           & 98.74           \\
LDAM Loss \cite{cao2019learning-ldam}   & 99.34           & 98.77           & 97.75           \\
PBS-SCL \cite{peng2022progressively}    & 99.84           & 99.79           & 99.62           \\
SCCAM        & \textbf{100.00} & \textbf{100.00} & \textbf{100.00} \\ \hline
\end{tabular}}
\label{table_csth_accuracy}
\end{table}

\subsubsection{Experimental Settings}

Originally, 1000 normal samples and 500 fault samples are generated with the procedure described in \cite{arunthavanathan2022autonomous-csthex}. Then the time-series data is segmented using the sliding-window method and the processed samples are divided into a training set and a test set. The details of the three settings are demonstrated in Table \ref{tabel_csth_setting}. %More specifically, for the balanced fault diagnosis, there are 780 normal samples and 450 fault samples in the training set while the test set contains 200 normal samples and 200 fault samples. On the other hand, for the imbalanced and long-tail fault diagnosis, only 200 and 20 fault samples are collected in the training sets respectively for limited fault diagnosis while the test sets are unchanged.

\subsubsection{Fault Diagnosis Results}

The specific fault diagnosis results of the proposed SCCAM method and the other four comparison methods are shown in Table \ref{table_csth_accuracy}. It can be seen that the proposed SCCAM method achieves 100.00\% fault detection accuracy in the three conditions, which is the best overall effect among all methods. %For instance, the average accuracy of the proposed SCCAM method is increased by 2.04\%, 1.28\%, 2.30\%, and 0.38\% compared to the focal loss, seesaw loss, LDAM loss, and PBS-SCL respectively for the long-tail fault diagnosis. 
These results prove the effectiveness of the proposed SCCAM method on fault classification under limited fault samples.

\subsubsection{Root Cause Analysis}

\begin{figure}[]
\centering
\subfloat[]
{
 \centering
 \includegraphics[width=4cm]{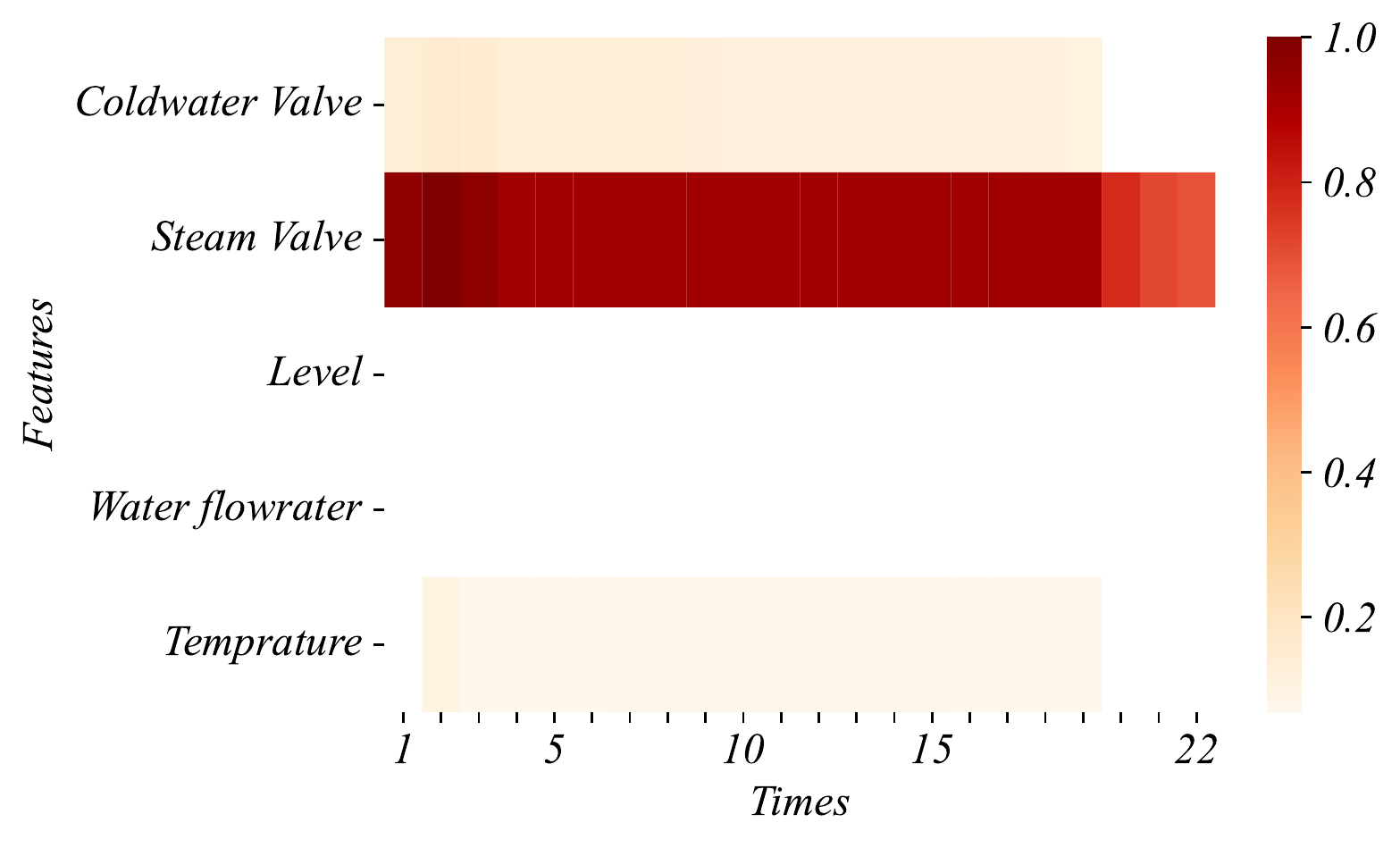}
 \label{fig_csth_global}
}
\hfill
\subfloat[]
{
 \centering
 \includegraphics[width=4cm]{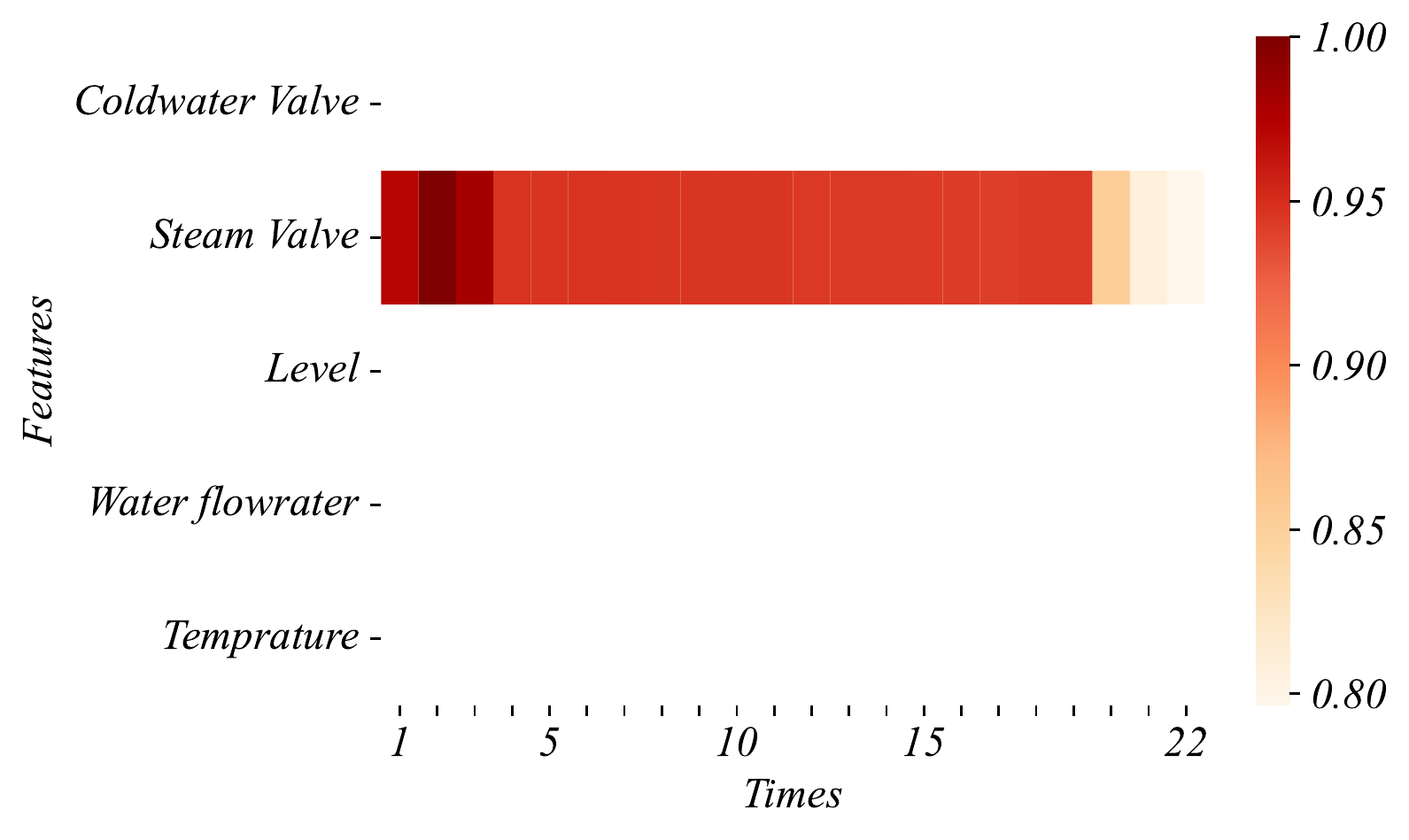}
 \label{fig_csth_local}
}
\caption{The visualization of (a) global and (b) local attention maps of fault on CSTH dataset.} 
\label{fig_csth}
\end{figure}

We then analyze the feature contribution and obtain the root causes based on CBAM in the long-tail scenario. For Fault 1, an operational deviation occurs in the steam valve position thus directly affecting the steam value \cite{arunthavanathan2022autonomous-csthex}. First, the feature contribution of the overall test set is shown to identify the root causes (\emph{i.e.}, global explanation). We compute the average attention map of the fault samples belonging to the Fault 1 class as shown in Fig. \ref{fig_csth_global}. It is obvious that the steam value is assigned the highest weight by the trained model, which is exactly the true root cause. Furthermore, we randomly select a single fault sample to demonstrate its feature contribution (\emph{i.e.}, local explanation) as shown in Fig. \ref{fig_csth_local}. The corresponding attention map is generated from the selected fault sample and the steam value is also considered as the root cause. In a word, the proposed SCCAM method can identify the root causes precisely by analyzing the feature maps generated by the CBAM module.

Since the CSTH dataset is relatively simple, we use the TE dataset to ulteriorly evaluate the performance of the proposed SCCAM method, which is more complex and challenging.

\begin{comment}
\begin{figure}[]
\centering
\subfloat[]
{
 \centering
 \includegraphics[width=4cm]{csth_few_shot_feature_importances.pdf}
}
\hfill
\subfloat[]
{
 \centering
 \includegraphics[width=4cm]{csth_confusion.pdf}
}
\caption{(a) The average of feature importance of the fault and (b) the confusion matrix on CSTH dataset. \textcolor{red}{need to be considered}} 
\label{fig_csth_fe_con}
\end{figure}
\end{comment}

\subsection{Case 2: Tennessee Eastman Process}

\subsubsection{Brief Introduction of TE dataset}

\begin{comment}
\begin{figure}[b]
\centering
\includegraphics[width = \linewidth]{te_flow.pdf}
\caption{Process flowchart of the TE process \cite{ma2020novel-teimg}.}
\label{fig_te_arch}
\end{figure}
\end{comment}

The TE process is simulated as a real industrial process and is widely used as a benchmark for fault diagnosis. %We apply the TE dataset to further evaluate the performance of the proposed SCCAM method as it is more complex and challenging than the CSTH dataset. 
The simulation model can be downloaded from the website: \url{http://depts.washington.edu/control/LARRY/TE/download.html}. %And the overall process is composed of five components and the flowchart is shown in Fig. \ref{fig_te_arch}. 
And the overall process consists of 22 continuous measured variables (X1-X22) and a total of 22 conditions as shown in Table \ref{table_te_fault}. We randomly select Fault 1, Fault 2, Fault 3, Fault 8, Fault 10, Fault 11, Fault 12, Fault 13, Fault 14, and Fault 20 combined with the normal case for fault diagnosis under limited fault samples. In addition, we use Fault 10 and Fault 11 for root cause analysis as researchers have explored their true root causes, which are X18 and X9 respectively \cite{yu2015nonlinear-te1011}.

\begin{table}[]
\centering
\caption{Process Scenarios for TE Dataset \cite{peng2022open}}
\begin{tabular}{cc}
\hline
Fault ID & Description                                                   \\ \hline
0        & Normal case                                                   \\
1        & A/C feed ratio, B composition constant (stream 4) step        \\
2        & B composition, A/C ratio constant (stream 4) step             \\
3        & D feed temperature (stream 2) step                            \\
4        & Reactor cooling water inlet temperature step                  \\
5        & Condenser cooling water inlet temperature step                \\
6        & A feed loss (stream 1) step                                   \\
7        & C header pressure loss - reduced availability (stream 4) step \\
8        & A, B, C feed composition (stream 4) random variation          \\
9        & D feed temperature (stream 2) random variation                \\
10       & C feed temperature (stream 4) random variation                \\
11       & Reactor cooling water inlet temperature random variation      \\
12       & Condenser cooling water inlet temperature random variation    \\
13       & Reaction kinetics slow drift                                  \\
14       & Reactor cooling water valve sticking                          \\
15       & Condenser cooling water valve sticking                        \\
16-20    & Unknown                                                       \\
21       & Valve for stream 4 fixed at the steady-state position         \\ \hline
\end{tabular}
\label{table_te_fault}
\end{table}

\begin{table}[b]
\caption{Experimental Settings on TE Dataset}
\centering
\setlength{\tabcolsep}{2.7mm}{
\begin{tabular}{ccccccc}
\hline
\multirow{2}{*}{Dataset} & \multicolumn{2}{c}{Balance} & \multicolumn{2}{c}{Imbalance} & \multicolumn{2}{c}{Long-tail} \\ \cline{2-7} 
                         & Normal        & Fault       & Normal         & Fault        & Normal        & Fault        \\ \hline
Train                    & 4780          & 4780        & 4780           & 478          & 4780          & 20           \\
Test                     & 780           & 780         & 780            & 780          & 780           & 780          \\ \hline
\end{tabular}}
\label{tabel_te_setting}
\end{table}

\subsubsection{Experiment Settings}

Motivated by the current data-driven fault diagnosis methods, we similarly use the downloaded simulation model to generate more data samples for evaluation \cite{song2022multi-data, he2021multiblock-data, peng2020cost-tedata}. By setting the sampling period to 36 seconds (100 samples/h), we generate 4800 training samples and 800 test samples for each class based on the simulation method from \cite{peng2020cost-tedata} on MATLAB. Then these time-series samples are then segmented and divided into the training set and the testing set. The setting details of the three conditions are indicated in Table. \ref{tabel_te_setting}. %And the overall dataset consists of a normal class and 10 fault classes. For the balanced fault diagnosis, the training set involves 4780 samples for each class while the test set contains 780 samples for each class. And for imbalanced and long-tail fault diagnosis, only 478 and 20 training samples are considered respectively for each fault class while the test sets are unchanged.

\subsubsection{Fault Diagnosis Results}

\begin{table}[]
\caption{Comparison of Fault Diagnosis Results on TE Dataset}
\centering
\setlength{\tabcolsep}{4.6mm}{
\begin{tabular}{cccc}
\hline
Method      & Balance         & Imbalance       & Long-tail       \\ \hline
Focal Loss \cite{lin2017focal}  & 97.32           & 91.07           & 65.45           \\
Seesaw Loss \cite{wang2021seesaw} & 97.06           & 90.86           & 62.83           \\
LDAM Loss \cite{cao2019learning-ldam}   & 97.15           & 91.80           & 65.15           \\
PBS-SCL \cite{peng2022progressively}    & 97.42           & 91.08           & 54.58           \\
SCCAM        & \textbf{98.00} & \textbf{94.70} & \textbf{72.16} \\ \hline
\end{tabular}}
\label{table_te_sota_accuracy}
\end{table}

The comparison results of the proposed SCCAM method are shown in Table. \ref{table_te_sota_accuracy}. It is obvious that the proposed SCCAM method outperforms the other four state-of-the-art long-tail learning algorithms and achieves the highest fault detection accuracy in all three conditions. The average accuracy of the proposed SCCAM method is 98.00\%, 94.70\%, and 72.16\% under balanced scenario, imbalanced scenario, and long-tail scenario respectively.
%increased by 10.25\%, 14.85\%, 10.76\%, and 32.21\% compared to the focal loss, seesaw loss, LDAM loss, and PBS-SCL respectively for the long-tail fault diagnosis.

\subsubsection{Root Cause Analysis}

We also evaluate the root causes identification ability of the proposed SCCAM method in the long-tail condition. Fault 10 and Fault 11 are selected for root cause analysis and the corresponding true root causes are X18 and X9. For Fault 10, the temperature in feed C randomly varies thus directly influencing the temperature of the stripper column (X18). And for Fault 11, the reactor temperature (X9) is affected by the random variation in reactor cooling water inlet temperature \cite{yu2015nonlinear-te1011}.

\begin{figure}[b]
\centering
\subfloat[]
{
 \centering
 \includegraphics[width=4cm]{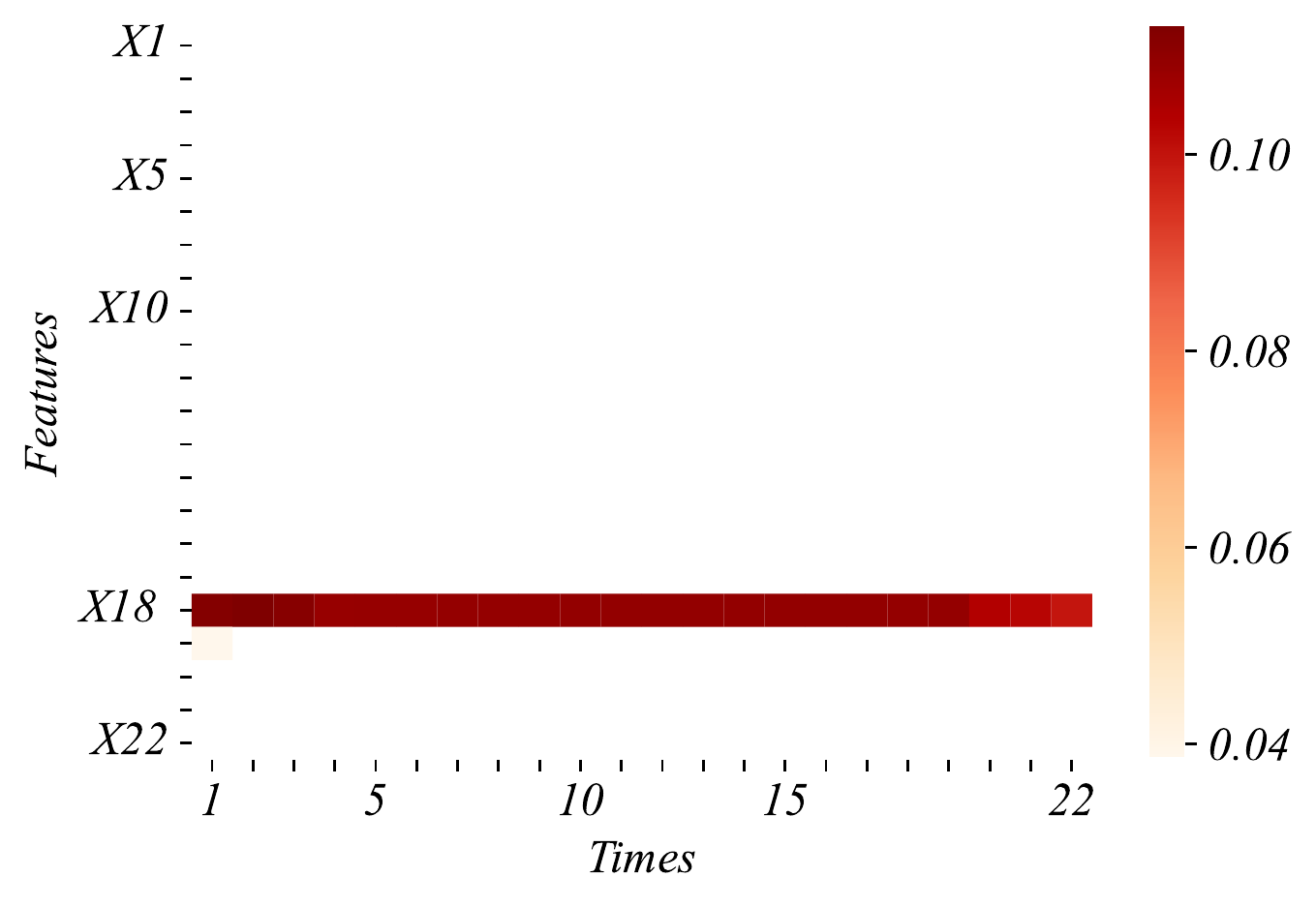}
}
\hfill
\subfloat[]
{
 \centering
 \includegraphics[width=4cm]{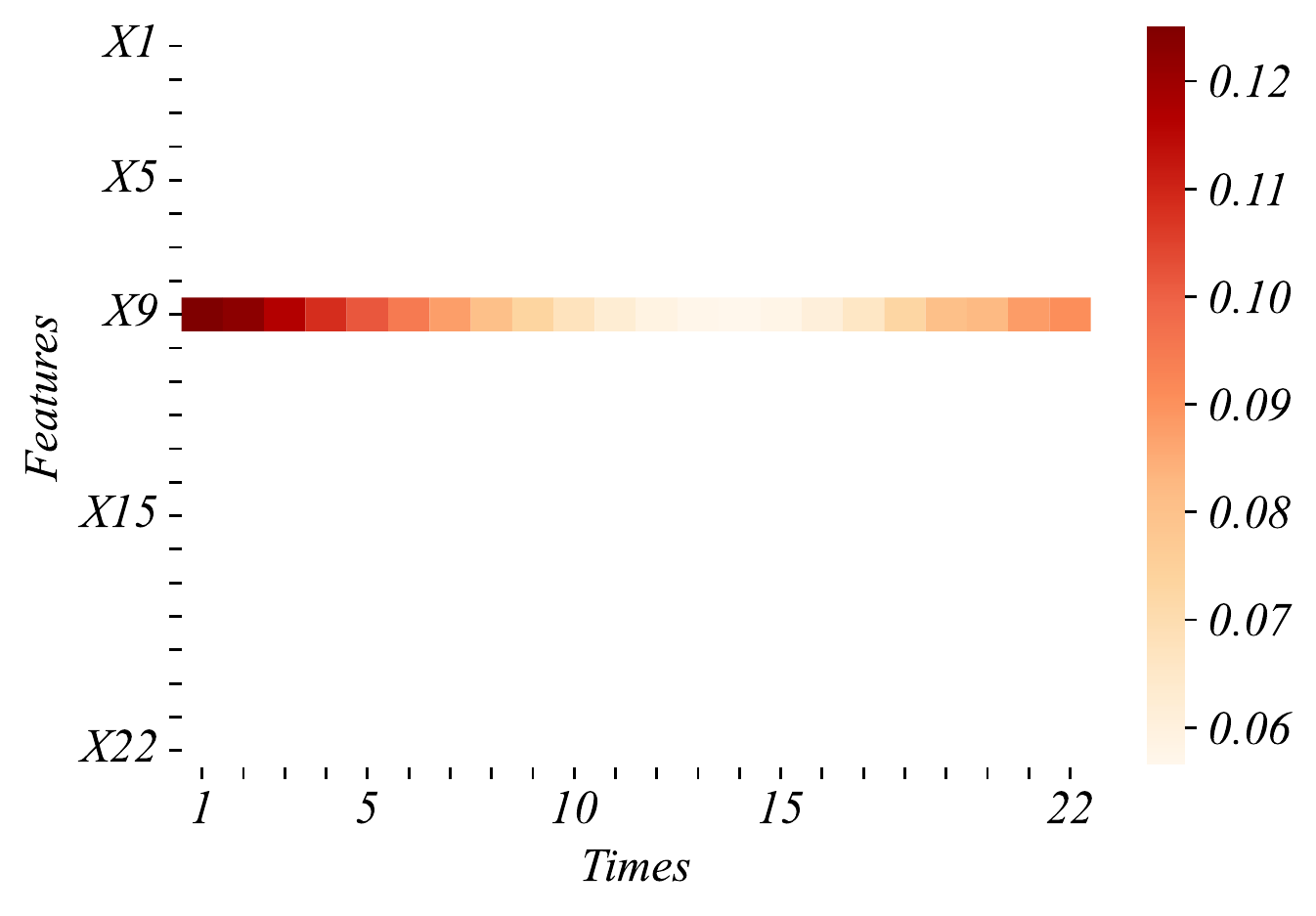}
}
\caption{The visualization of global attention maps generated by our SCCAM proposed method of (a) Fault 10 and (b) Fault 11 on TE dataset.} 
\label{te_global_1}
\end{figure}

\begin{figure}[]
\centering
\subfloat[]
{
 \centering
 \includegraphics[width=4cm]{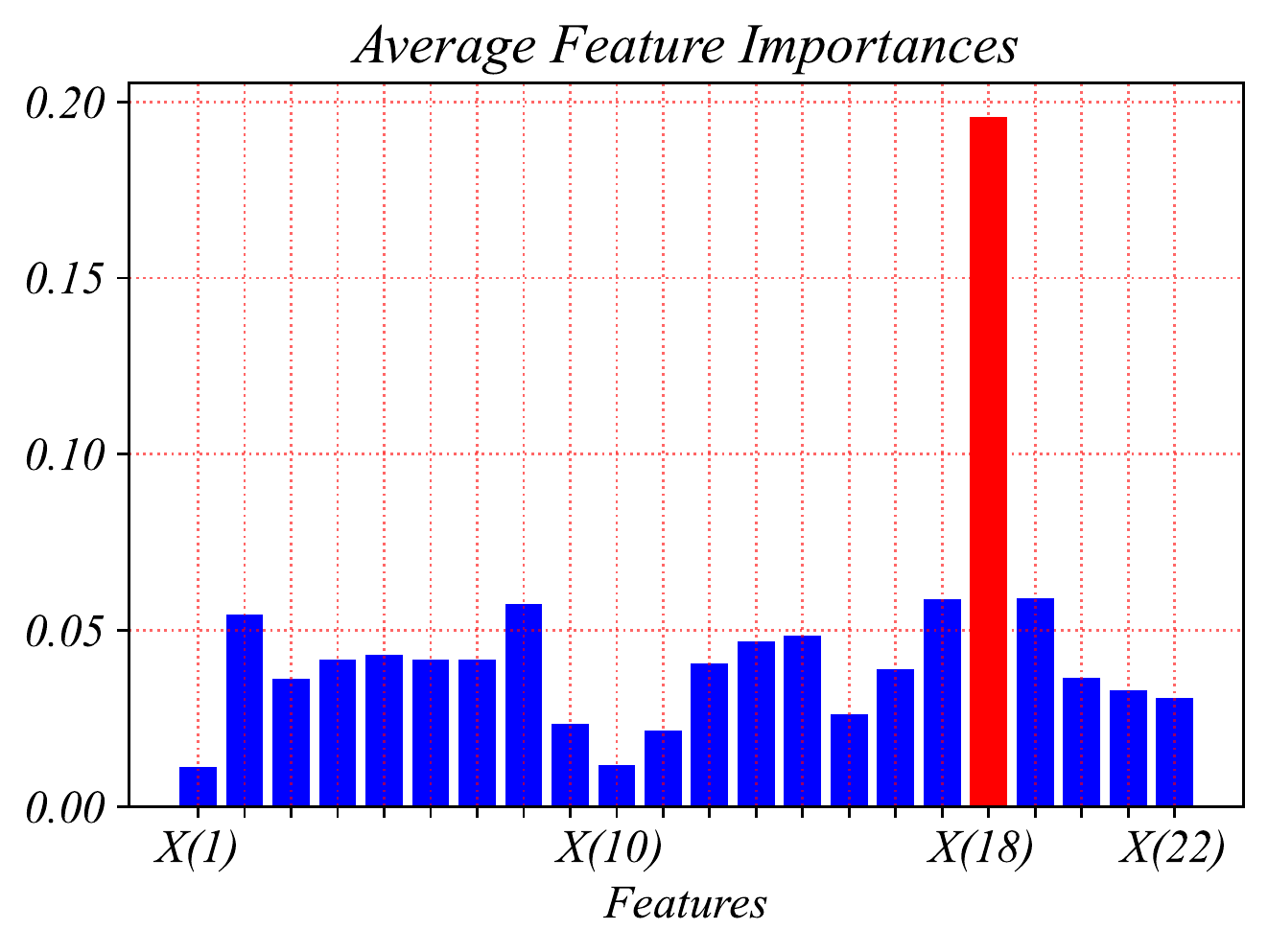}
}
\hfill
\subfloat[]
{
 \centering
 \includegraphics[width=4cm]{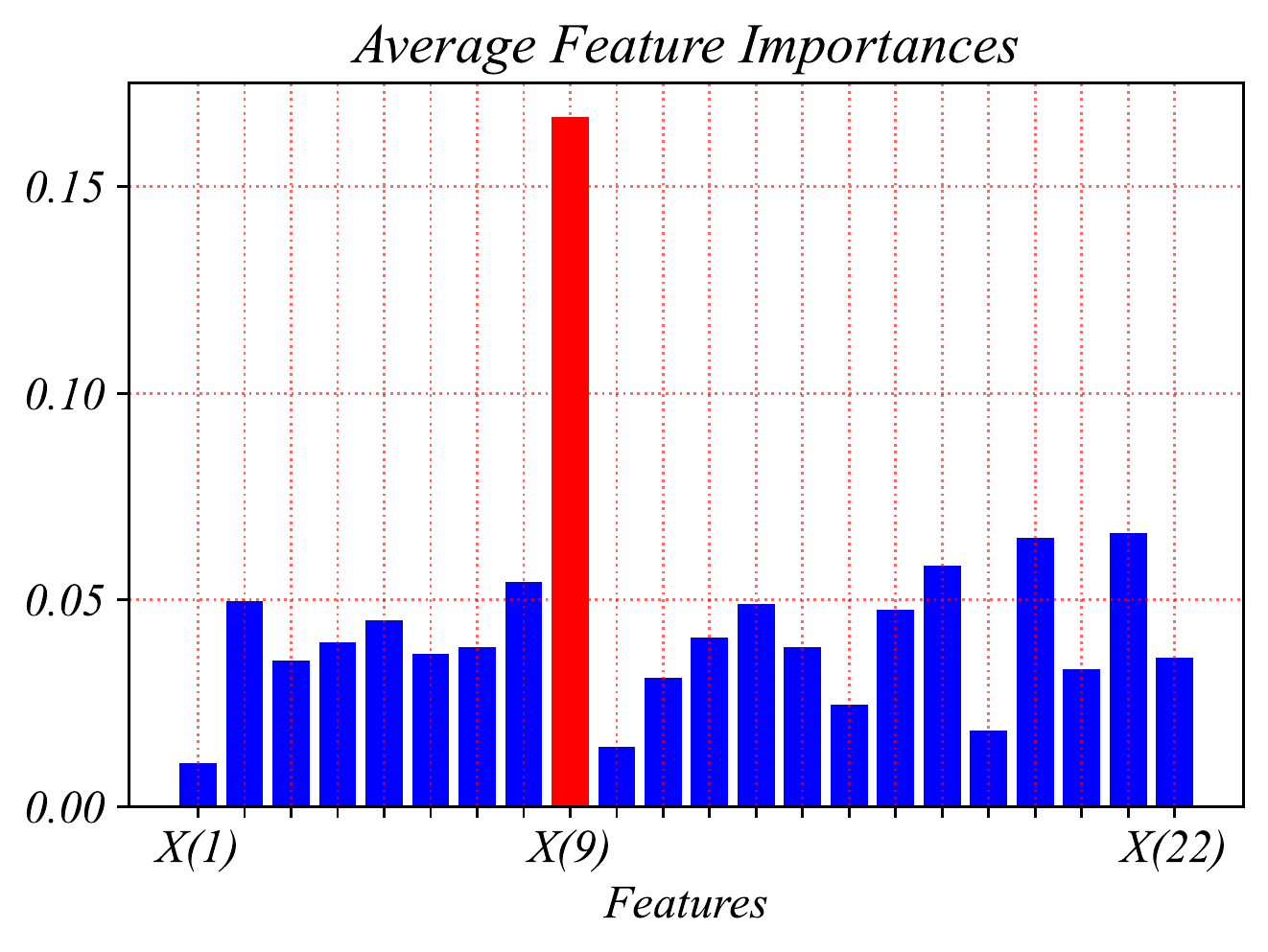}
}
\caption{The average feature importance generated by our proposed SCCAM method of (a) Fault 10 and (b) Fault 11 on TE dataset.} 
\label{te_feature_importance}
\end{figure}

First, we calculate the average attention map of the fault samples belonging to the same fault type (\emph{i.e.}, global explanations). The obtained heatmaps are shown in Fig. \ref{te_global_1}. We see that the variables X18 and X9 are assigned with the highest feature contribution respectively, which are consistent with the true root causes. Then we squeeze the time domain to compute the average feature importance under extremely limited fault samples. As shown in Fig. \ref{te_feature_importance}, X18 and X9 are considered as the most important features of Fault 10 and 11 respectively. 

\begin{figure}[]
\centering
\subfloat[]
{
 \centering
 \includegraphics[width=4cm]{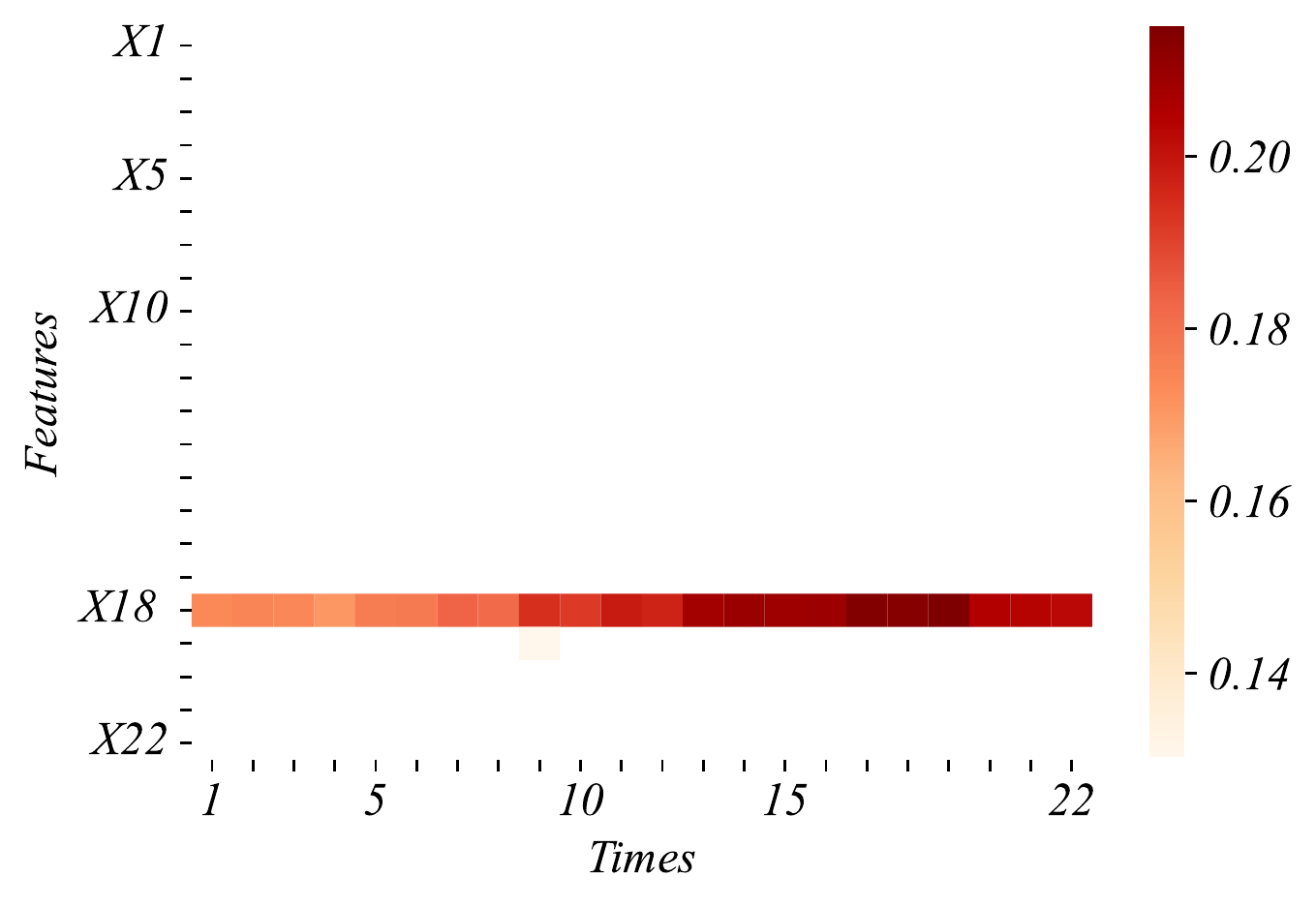}
}
\hfill
\subfloat[]
{
 \centering
 \includegraphics[width=4cm]{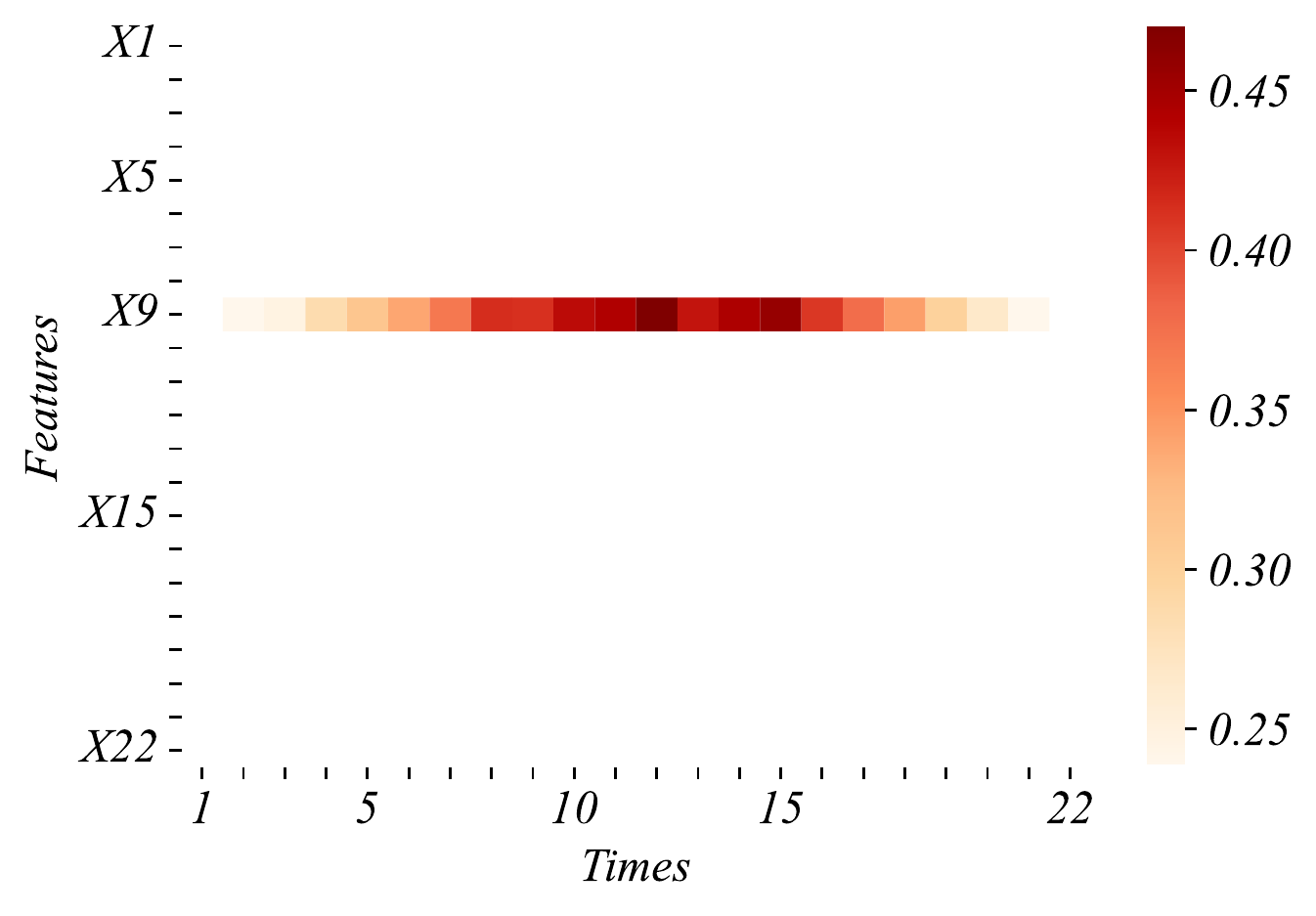}
}
\caption{The visualization of local attention maps generated by our SCCAM proposed method of (a) Fault 10 and (b) Fault 11 on TE dataset.} 
\label{te_local}
\end{figure}

Then the feature contribution from a single fault sample is demonstrated (\emph{i.e.}, local explanations). We randomly select a sample from Fault 10 and compute its attention map by applying a channel-wise average pooling operation. Similarly, the attention map of Fault 11 is obtained. As shown in Fig. \ref{te_local}, X18 and X9 refer to the most important features for Fault 10 and 11 respectively. It proves that the proposed SCCAM method has the ability to identify the root causes even under extremely limited fault samples. In other words, the abnormal variation in the industrial process can be correctly captured by the proposed SCCAM method.

\subsubsection{Inductive Biases Discussion}

\begin{figure}[]
\centering
\includegraphics[width = \linewidth]{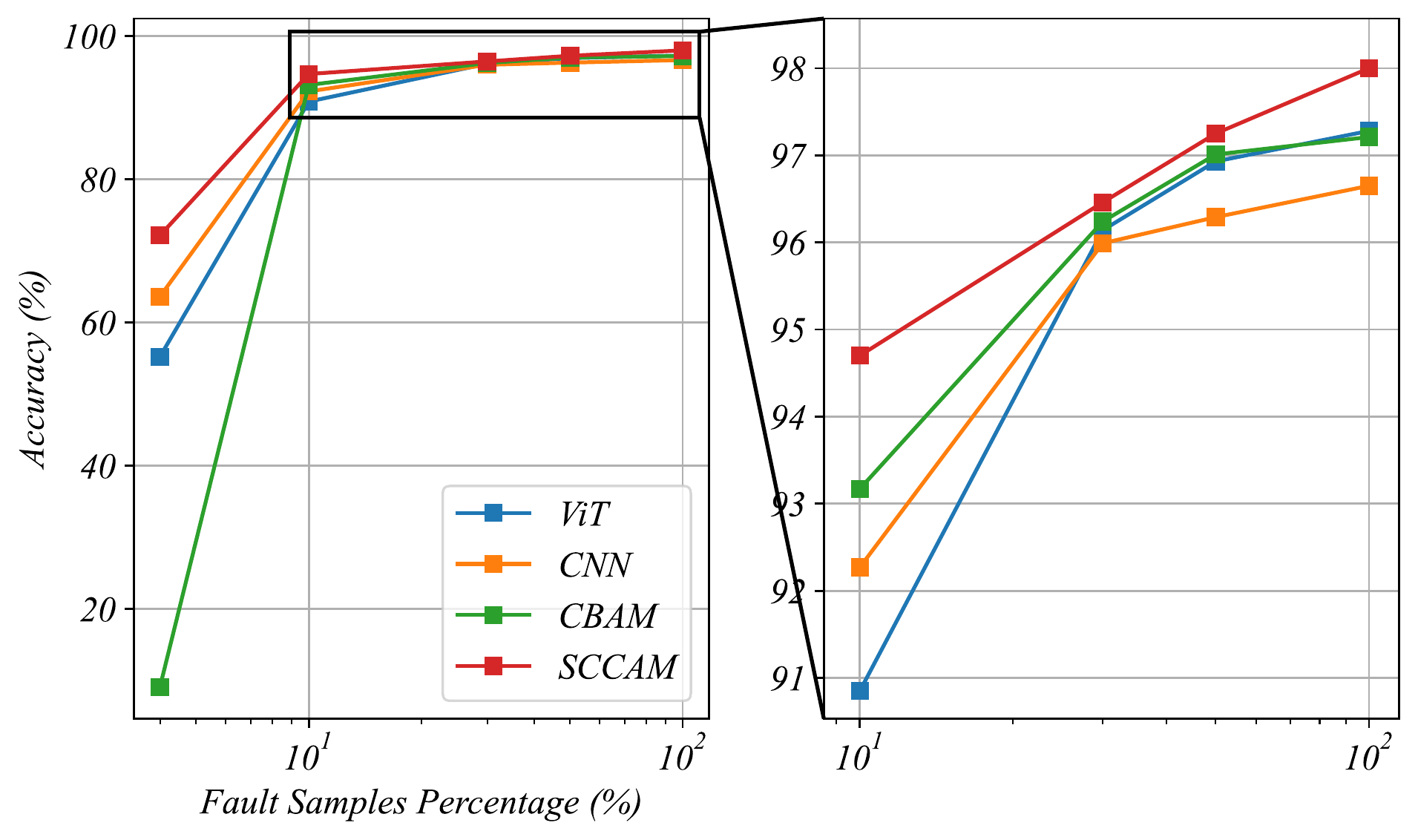}
\caption{Comparisons of the sample efficiency on TE dataset.}
\label{fig_sample_efficiency}
\end{figure}

To ulteriorly explain the impact of inductive biases on model performance, we conduct supplementary experiments and compare the sample efficiency. Three comparison methods are considered, including vision transformer (ViT) \cite{dosovitskiy2020image-vit}, CNN \cite{wang2020intelligent-tcy}, and CBAM \cite{woo2018cbam}. As shown in Fig. \ref{fig_sample_efficiency}, our SCCAM method outperforms the other three methods. With enough fault data (\emph{i.e.}, 100\%, 50\%, and 30\% fault samples), ViT and CBAM achieve similar performance while CNN has the worst effect. This is consistent with the fact that attention-based models have better generalization capability on datasets with relatively enough fault samples. As the fault samples percentage is further reduced, we obtain different results. When trained with fewer fault samples (\emph{i.e.}, 10\% fault samples), ViT has the worst effect compared with CNN and CBAM. The reason why this happens is that ViT lacks intrinsic inductive biases thereby performing worse in a limited data setting. And in the long-tail scenario (\emph{i.e.}, 4\% fault samples), CNN achieves better performance than ViT due to its strong intrinsic inductive biases of locality and spatial invariance. However, CBAM randomly classifies the fault samples and has the worst effect. This is because directly integrating attention mechanisms with CNN increases the number of parameters and complexity of the original model thereby causing the overfitting problem under extremely limited fault samples. And our SCCAM method avoids this issue by utilizing SCL loss to efficiently leverage each limited fault sample and learn powerful feature representations.

\begin{figure*}[b]
\centering
\subfloat[]
{
 \centering
 \includegraphics[width=4cm]{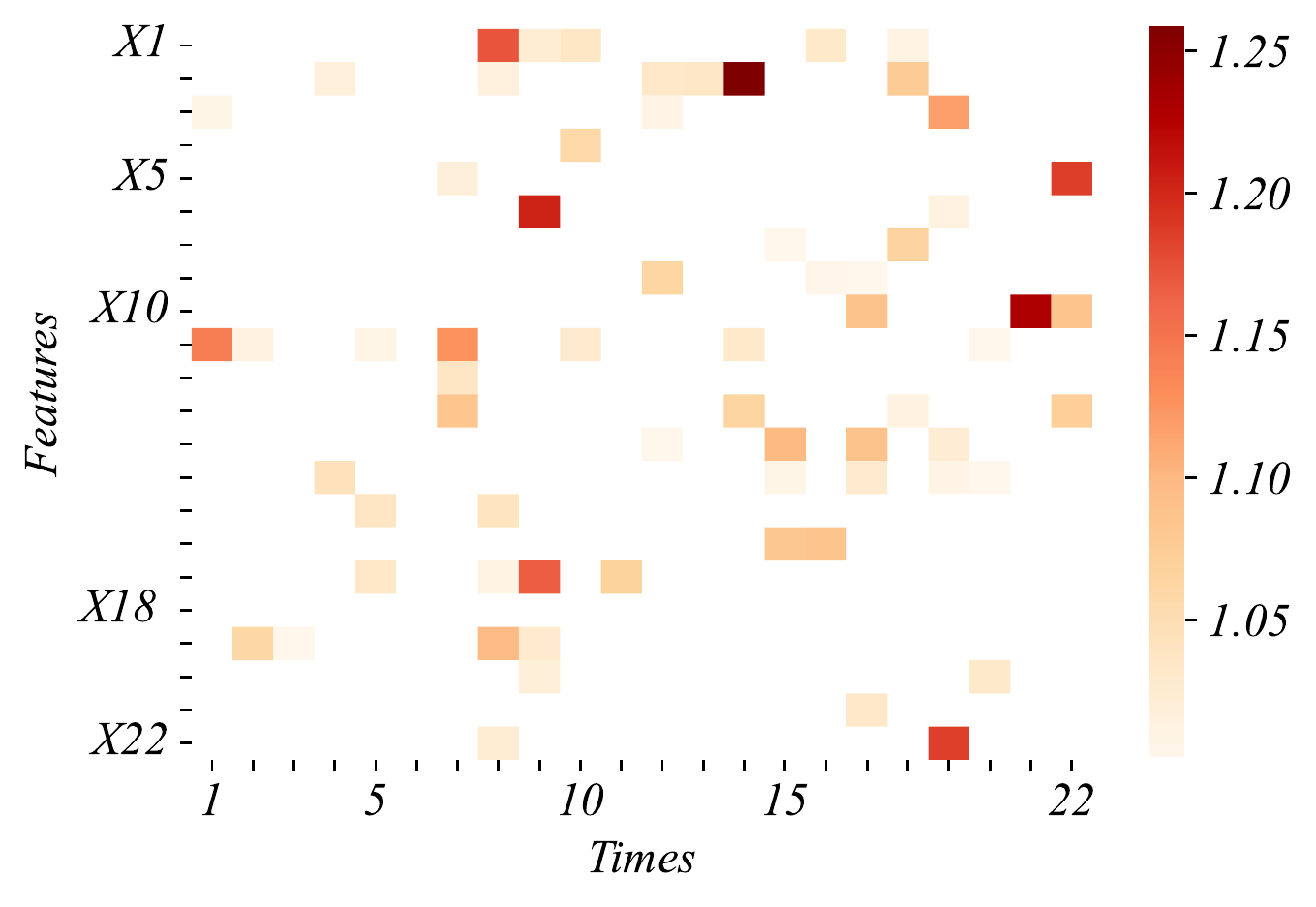}
}
\hfill
\subfloat[]
{
 \centering
 \includegraphics[width=4cm]{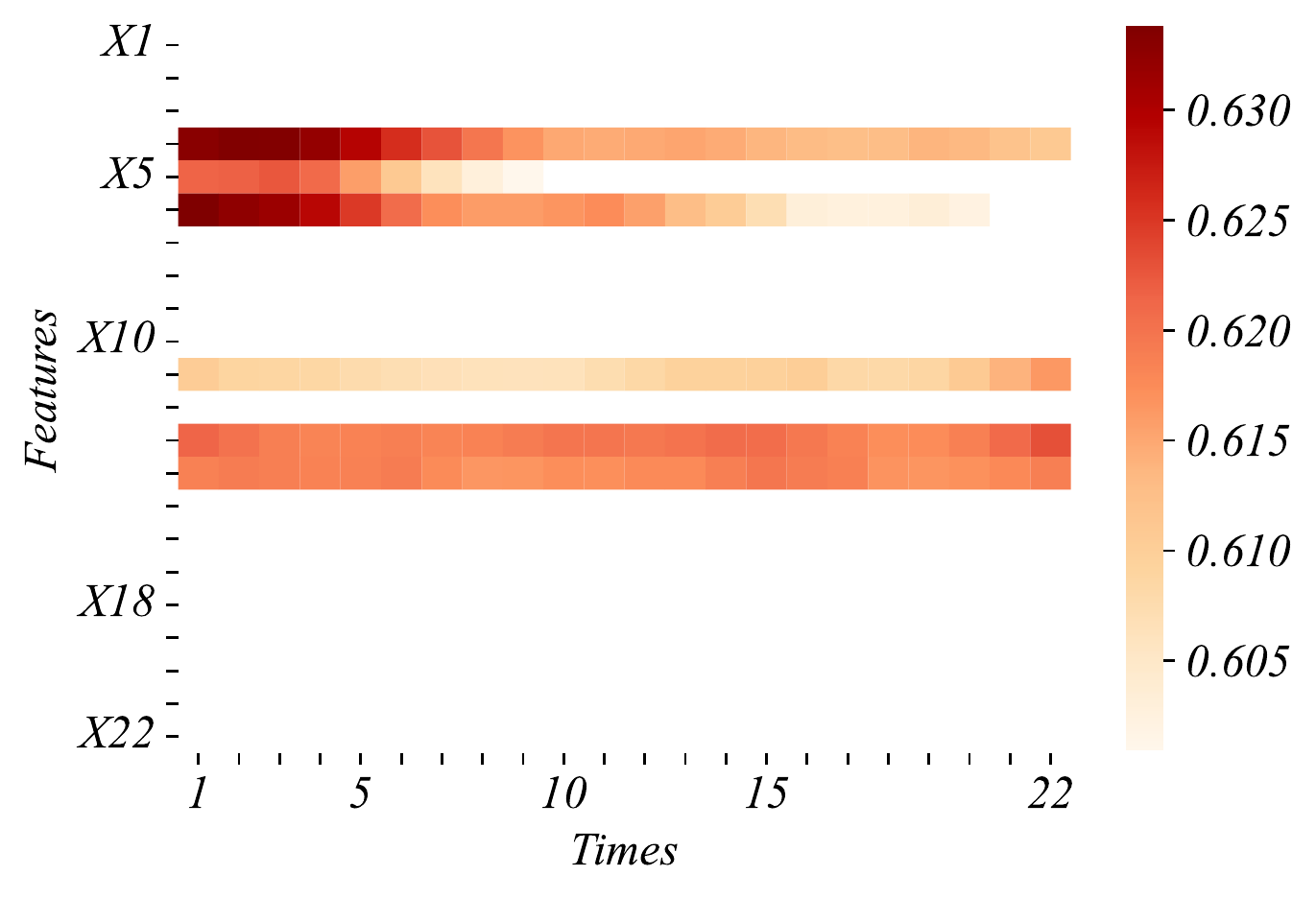}
}
\hfill
\subfloat[]
{
 \centering
 \includegraphics[width=4cm]{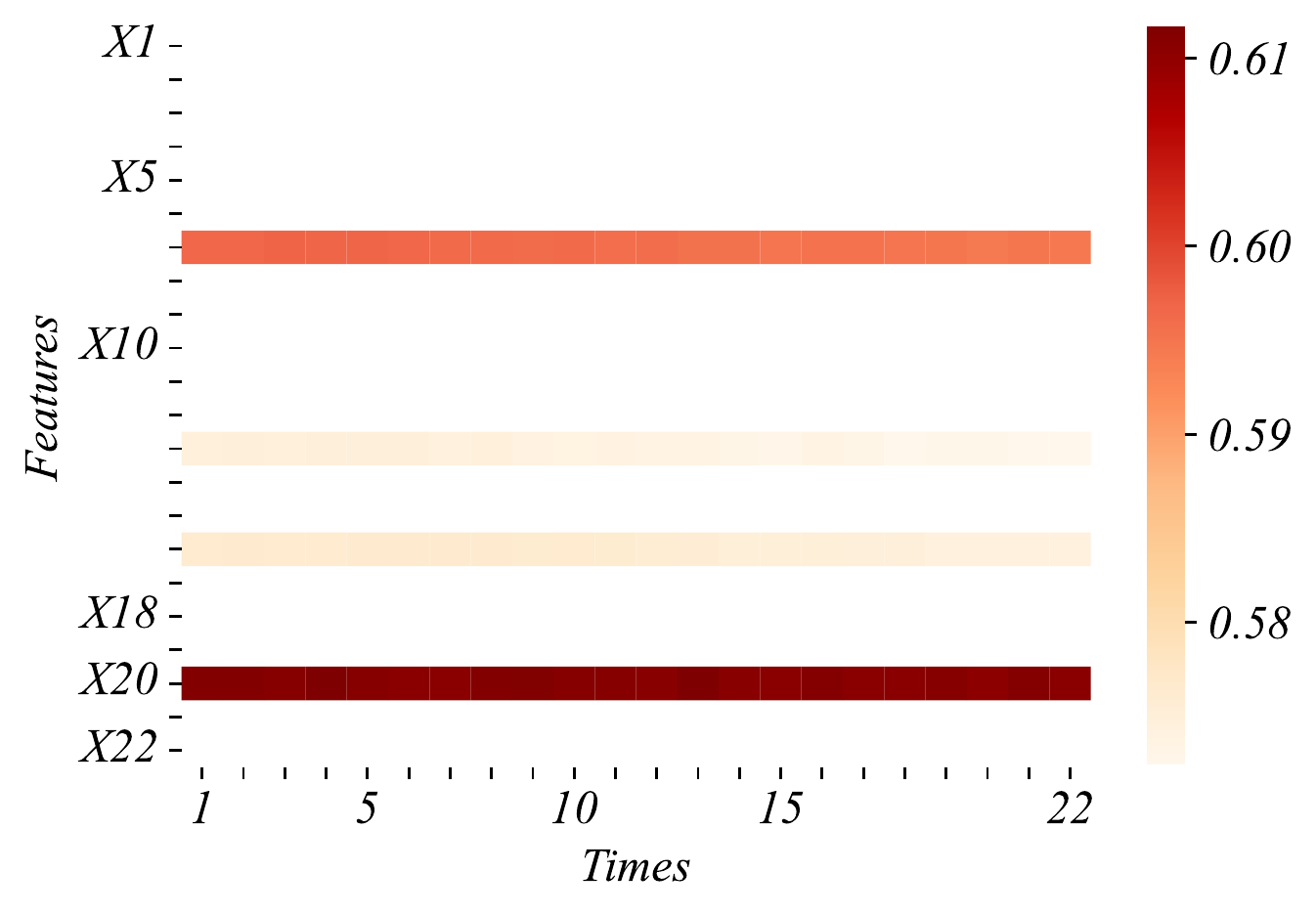}
}
\hfill
\subfloat[]
{
 \centering
 \includegraphics[width=4cm]{te_few_shot_global_class10.pdf}
}
\caption{The visualization of global attention maps of Fault 10 generated by (a) model A, (b) model B, (c) model D, and (d) SCCAM method on TE dataset.} 
\label{te_global_10}
\end{figure*}

\begin{figure*}[b]
\centering
\subfloat[]
{
 \centering
 \includegraphics[width=4cm]{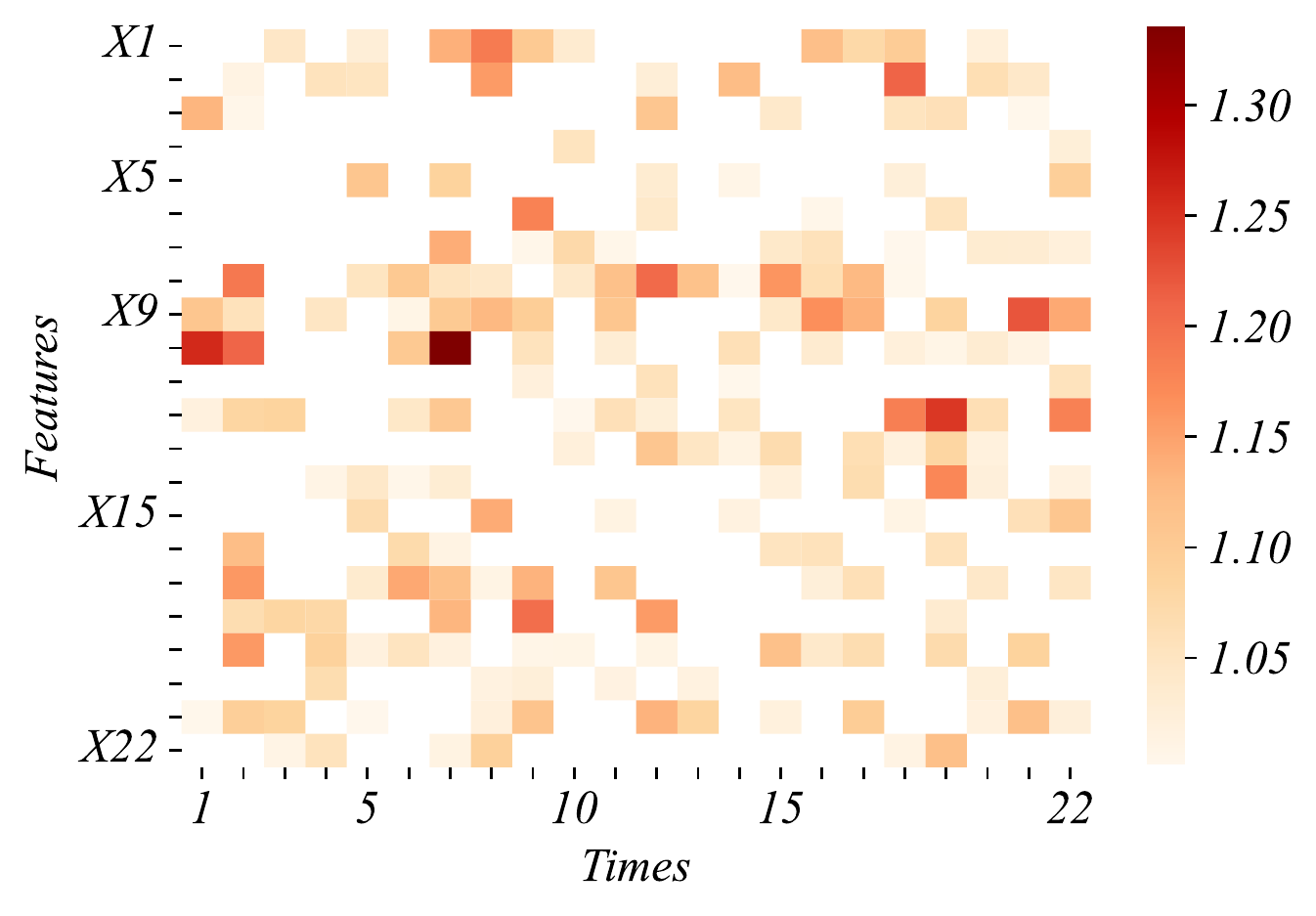}
}
\hfill
\subfloat[]
{
 \centering
 \includegraphics[width=4cm]{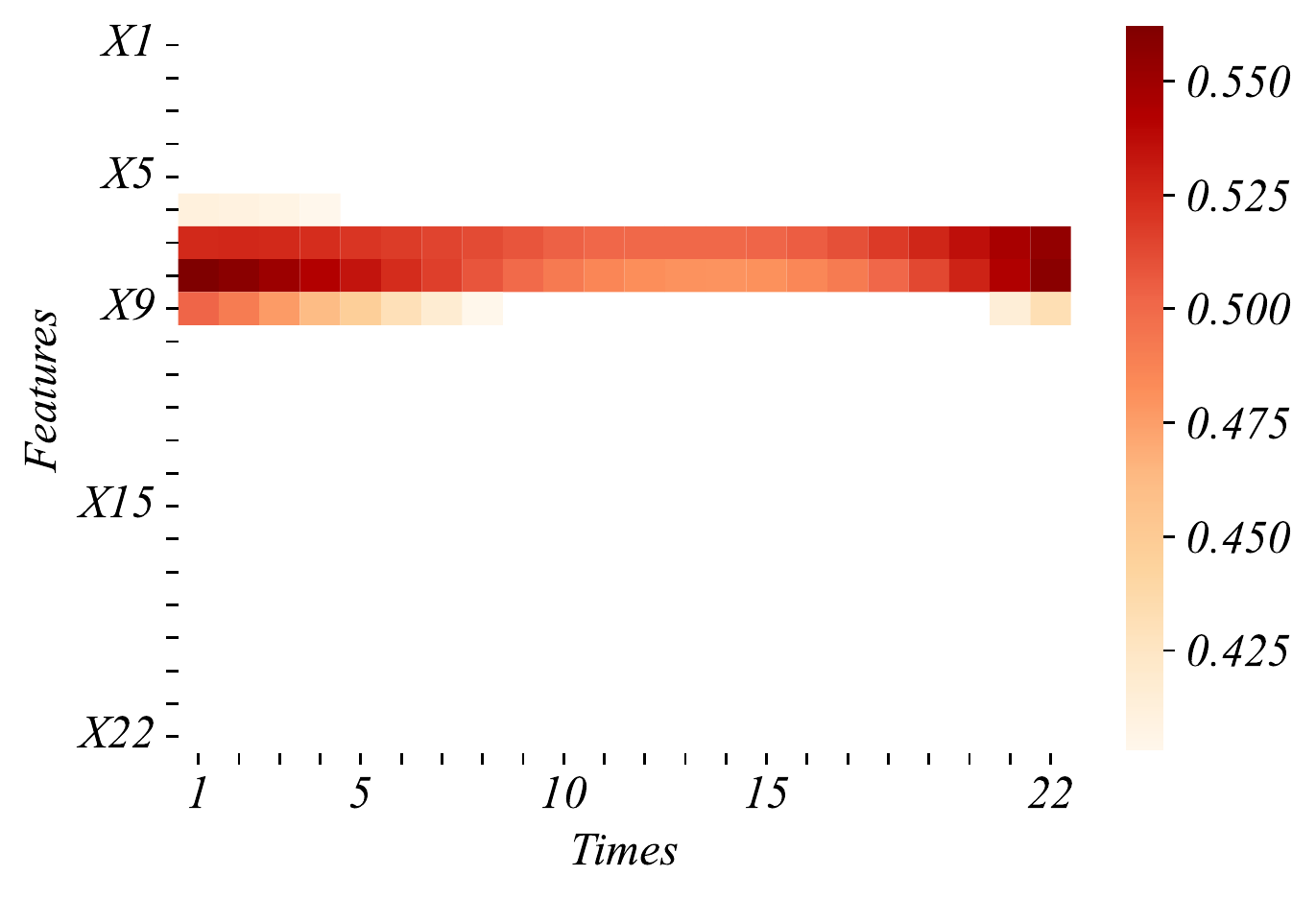}
}
\hfill
\subfloat[]
{
 \centering
 \includegraphics[width=4cm]{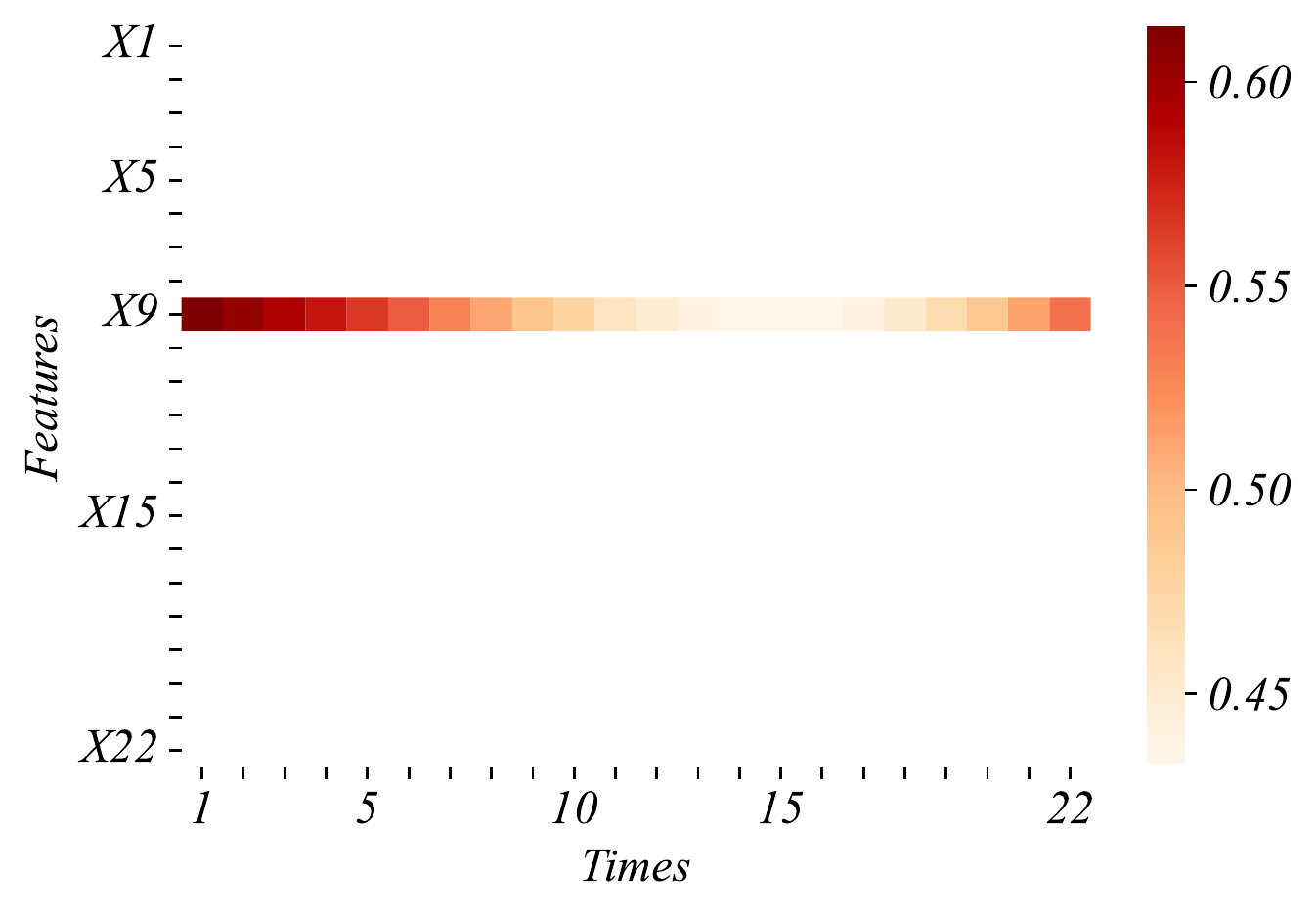}
}
\hfill
\subfloat[]
{
 \centering
 \includegraphics[width=4cm]{te_few_shot_global_class11.pdf}
}
\caption{The visualization of global attention maps of Fault 11 generated by (a) model A, (b) model B, (c) model D, and (d) SCCAM method on TE dataset.} 
\label{te_global_11}
\end{figure*}

\subsubsection{Ablation Experiments}

% \usepackage{multirow}
\begin{table}[]
\caption{Ablation Experiments of Fault Diagnosis on TE Dataset}
\centering
\setlength{\tabcolsep}{1.25mm}{
\begin{tabular}{cccccccc}
\hline
\multirow{2}{*}{Method} & \multirow{2}{*}{Attention} & \multirow{2}{*}{CNN} & \multirow{2}{*}{\begin{tabular}[c]{@{}c@{}}CE\\ Loss\end{tabular}} & \multirow{2}{*}{\begin{tabular}[c]{@{}c@{}}SCL\\ Loss\end{tabular}} & \multicolumn{3}{c}{Average Accuracy (\%)}        \\ \cline{6-8} 
                        &                                                                        &                      &                                                                    &                                                                     & Balance        & Imbalance      & Long-tail      \\ \hline
A                     & \checkmark                                                                      &                      & \checkmark                                                                  &                                                                     & 97.28          & 90.85          & 55.22          \\
B                     &                                                                        & \checkmark                    & \checkmark                                                                  &                                                                     & 96.65          & 92.27          & 63.62          \\
C               & \checkmark                                                                      & \checkmark                    & \checkmark                                                                  &                                                                     & 97.21          & 93.17          & 9.09           \\
SCCAM                    & \checkmark                                                                      & \checkmark                    &                                                                    & \checkmark                                                                   & \textbf{98.00} & \textbf{94.70} & \textbf{72.16} \\ \hline
\end{tabular}}
\label{table_te_accuracy}
\end{table}

In this section, detailed ablation experiments are performed for both fault classification and root cause analysis. First, the ablation experiments for fault classification are shown in Table. \ref{table_te_accuracy}. Our proposed SCCAM method achieves the best performance in all three settings. In the balanced scenario, all four methods can accurately classify the faults since enough training samples are provided to extract feature information. And in the imbalanced scenario, the fault detection accuracy of model A, model B, model C, and our proposed model is reduced by 6.61\%, 4.53\%, 4.16\%, and 3.37\% respectively, due to the reduction of the fault samples. It is obvious that the proposed SCCAM method is the least affected. Although it seems that model A performs well, things change in the long-tail condition. For long-tail diagnosis, model A has worse performance compared to model B and fails to classify the faults with a fault detection accuracy of only 55.22\%. It is because the pure attention-based model A lacks intrinsic inductive bias thus requiring more data for training and performing worse under limited fault samples. In addition, model C randomly classifies the samples and has the worst effect with a fault detection accuracy of 9.09\% due to the overfitting problem. In contrast, the proposed SCCAM method achieves the best performance among the four methods. These results prove the effectiveness of the combination of attention mechanisms and CNN, and verify the feature learning capability of the SCL loss.
%More specifically, the fault detection accuracy of our proposed SCCAM method is 30.68\%, 13.42\%, and 693.84\% higher compared to the results of model A, model B, and model C respectively.

\begin{table}[]
\centering
\caption{Ablation Settings for Root Cause Analysis}
\setlength{\tabcolsep}{1.3mm}{
\begin{tabular}{cccccccc}
\hline
Method        & CNN & \begin{tabular}[c]{@{}c@{}}1*1\\ Kernel\end{tabular} & Grad-CAM & CBAM & \begin{tabular}[c]{@{}c@{}}Multi-head \\ Attention\end{tabular} & \begin{tabular}[c]{@{}c@{}}CE\\ Loss\end{tabular} & \begin{tabular}[c]{@{}c@{}}SCL\\ Loss\end{tabular} \\ \hline
A           &     &                                                      &          &      & \checkmark                                                               & \checkmark                                                 &                                                    \\
B             & \checkmark   &                                                     & \checkmark        &      &                                                                 & \checkmark                                                 &                                                    \\
D             & \checkmark   & \checkmark                                                     & \checkmark        &      &                                                                 & \checkmark                                                 &                                                    \\
SCCAM & \checkmark   & \checkmark                                                    &          & \checkmark    &                                                                 &                                                   & \checkmark                                                  \\ \hline
\end{tabular}}
\label{ablation_root}
\end{table}

Second, we generate global explanations in the long-tail scenario to analyze the ability to identify the root causes and further verify the effectiveness of the used $1 \times 1$ convolution kernels. The ablation settings are shown in Table. \ref{ablation_root}. We use the internal multi-head attention to show feature contribution for model A. In addition, model B involves the commonly used $3 \times 3$ kernels while model D has $1 \times 1$ kernels. Then the commonly used post-hoc interpretable gradient-weighted class activation mapping (Grad-CAM) \cite{selvaraju2017grad-cam} is applied to identify the root causes for model B and model D. The obtained heatmaps are shown in Fig. \ref{te_global_10} and Fig. \ref{te_global_11}. Only our proposed SCCAM method can identify the true root causes for both Fault 10 and Fault 11. Although model A can directly provide feature contribution through its internal attention maps, its multiple heads focus on various feature dimensions and fail to identify the true root causes stably. In addition, model B mislocates the root causes for both fault types while model D succeeds to find the true root cause for Fault 11. This proves the effectiveness of the $1 \times 1$ convolution kernels.

\section{Conclusion}
\label{conclusion}

This paper innovatively proposed SCCAM, an ante-hoc interpretable framework for fault diagnosis under limited fault samples. It solves the root cause analysis problem under limited fault samples for the first time. The core idea is to utilize CNN to incorporate intrinsic inductive biases into attention-based architectures thus improving the classification capability under limited fault samples. Meanwhile, the $1 \times 1$ convolution kernels are applied to enable feature-level explanations and enhance the interpretability of attention mechanisms. We evaluate the proposed SCCAM method on both CSTH and TE datasets. Three common fault diagnosis scenarios are involved: a balanced scenario for additional verification and two scenarios with limited fault samples (\emph{i.e.}, imbalanced scenario and long-tail scenario). Comprehensive experiments show that SCCAM outperforms other state-of-the-art methods in various respects including sample efficiency, fault classification, and root cause analysis.

In our future work, we plan to explore the open-set fault diagnosis. In this case, the classifier needs to handle samples belonging to unknown fault types. Although some achievements have been reported in this field recently, few studies have focused on root cause analysis for open-set fault diagnosis. The exploration of this challenging problem is necessary and significative.

\bibliographystyle{IEEEtran}%
\bibliography{ref.bib}
\vspace{12pt}

\end{document}